\documentclass[letterpaper]{article}

\usepackage{aaai2027}
\nocopyright

\usepackage[hyphens]{url}
\usepackage{graphicx}
\usepackage{natbib}
\usepackage{caption}
\usepackage{xcolor}

\definecolor{gtText}{RGB}{90,90,90}
\definecolor{mmgpeText}{RGB}{176,108,28}
\definecolor{regText}{RGB}{38,132,168}
\definecolor{clsText}{RGB}{124,84,158}

\usepackage{booktabs}
\usepackage{multirow}
\usepackage{makecell}
\usepackage{enumitem}
\usepackage{amsmath}
\usepackage{amssymb}
\usepackage{cleveref}
\usepackage{placeins}

\graphicspath{{Figures/}{figs/}}
\providecommand{\Description}[1]{}

\title{
mmSimPrior: Learning Simulation Priors for Data-Efficient
Real-World Generalizable Radar-Based Human Motion Reconstruction
}

\author{
Cheng Guo\textsuperscript{\rm 1},
Qiming Cao\textsuperscript{\rm 2},
Shengkai Xu\textsuperscript{\rm 1},\\
Haoyu Xie\textsuperscript{\rm 1},
Kaixiang Su\textsuperscript{\rm 1},
Pu Wang\textsuperscript{\rm 1},
Hongfei Xue\textsuperscript{\rm 1}\corresponding
}

\affiliations{
\textsuperscript{\rm 1}University of North Carolina at Charlotte, Charlotte, NC, USA\\
\textsuperscript{\rm 2}Purdue University, West Lafayette, IN, USA\\
\{cguo3, sxu7, hxie3, ksu4, pwang13, hxue2\}@charlotte.edu, cao393@purdue.edu
}

\begin{document}

\maketitle

% =========================================================
% Main paper
% =========================================================

\begin{abstract}
Millimeter-wave (mmWave) radar enables privacy-preserving and illumination-robust human motion reconstruction, but training generalizable models typically require costly paired radar–motion recordings. 
Simulation can scale such supervision, yet even physics-based simulators cannot fully reproduce real-world multipath, clutter, hardware-specific response statistics, or distance-dependent resolution degradation, leaving a sim-to-real gap. 
We present \textbf{mmSimPrior}, a simulation-pretrained framework that factorizes transferable knowledge into signal, motion, and radar-to-motion mapping priors.
To learn transferable signal and motion priors, we pretrain a multimodal radar encoder with a physics-informed domain-randomization curriculum designed to mitigate the sim-to-real gap by approximating real-world propagation- and acquisition-level variations, while a joint-temporal tokenizer learns a discrete prior over plausible human motion.
A dual-mode mapping module predicts either motion-code distributions for structurally constrained zero-shot reconstruction or continuous motion parameters for flexible adaptation from limited real data. 
We further construct a 4.2M-frame, 31K-sequence dataset suite and introduce a No-Overlap Setting that prevents any exact subject–environment–location–motion tuple from appearing in both the adaptation and test sets.
Experiments on mmSimPrior-Real and RT-Pose demonstrate consistent gains: with only 24 paired real sequences, mmSimPrior-Reg reduces MPJPE by 24.7-39.0\% over the strongest baseline across the three environments, while mmSimPrior-Cls reduces zero-shot MPJPE by 8.5\% without finetuning.
\end{abstract}

\begin{links}
\link{Project Page}{https://ch3ngguo.github.io/mmsimprior/}
\end{links}

\section{Introduction}
\label{sec:intro}

\begin{figure*}[t]
  \centering
  \includegraphics[width=\linewidth]{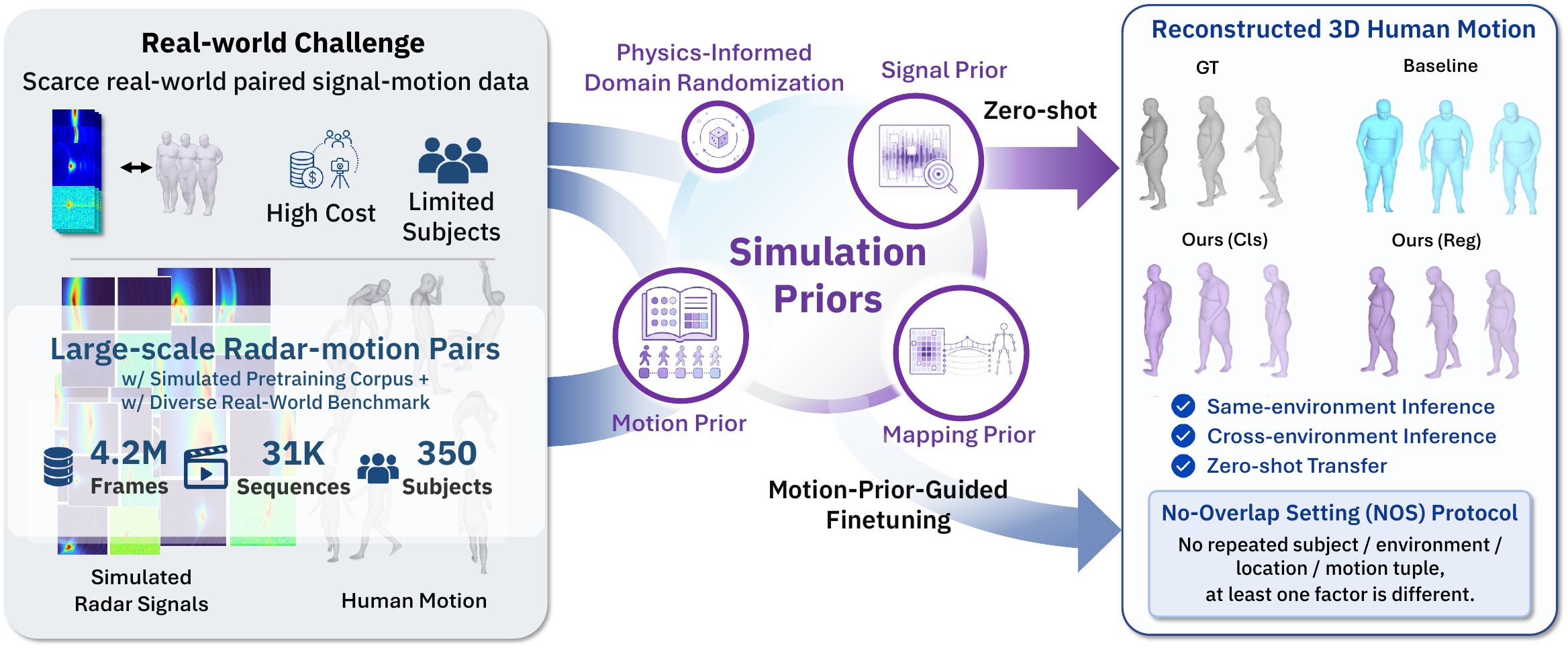}
  \caption{
    \textbf{Overview of mmSimPrior.}
    We construct large-scale simulated radar--motion pairs and learn
    transferable signal, motion, and mapping priors with physics-informed
    domain randomization.
    The learned priors support zero-shot transfer and motion-prior-guided
    real-world adaptation under a no-overlap evaluation protocol.
  }
  \label{fig:system_overview}
\end{figure*}

Human motion reconstruction supports applications in AR/VR, robotics,
and health monitoring~\cite{von2018recovering,goel2023humans,
chen2022mmbody,an2021mars}.
Although RGB and RGB-D methods have achieved substantial progress
~\cite{ionescu2013human3,mehta2017monocular,kanazawa2018hmr,
kolotouros2019spin,kocabas2020vibe,goel2023humans}, optical sensors
degrade under poor illumination and occlusion and raise privacy concerns
in sensitive indoor environments.
Millimeter-wave (mmWave) radar instead provides range, Doppler, and
angular measurements while remaining insensitive to lighting and
preserving visual privacy~\cite{richards2014radar,chen2011microdoppler,
zhao2018rf,zhao2018through,zhao2019through}.

Real-world radar-based human motion reconstruction remains challenging
because radar responses are sparse, noisy, and jointly determined by body
motion, subject geometry, sensing configuration, and the environment.
Without diverse paired coverage, models can mistake
deployment-specific reflection patterns for motion evidence.
However, collecting radar--motion annotations requires participant recruitment, sensor calibration, synchronization, and repeated recordings across environments, leaving existing datasets limited in scale and models prone to collection-specific overfitting.

Simulation offers scalable paired supervision and broad motion coverage
~\cite{zhang2022synthesized,xue2023towards,chen2023rf,deng2023midas,
joshi2024towards,huang2025one}, but scale alone does not ensure
real-world transfer.
Even physics-based simulators cannot fully reproduce multipath, clutter,
hardware-dependent response statistics, or distance-dependent resolution
degradation~\cite{richards2014radar,chen2011microdoppler,
xue2023towards,chen2023rf}. Exhaustive resimulation is costly, while
generic image augmentations may distort the range, Doppler, and spatial
semantics of radar heatmaps.
The central question is therefore whether simulation can serve as the
primary source of paired supervision for fine-grained radar-based
human motion reconstruction, rather than merely augmenting a
deployment-specific real training set.

We address this question with \textbf{mmSimPrior}, a simulation-pretrained
framework for zero-shot and limited-data real-world transfer.
It factorizes transferable knowledge into signal, motion, and mapping priors.
A multi-modal signal prior is pretrained with physics-informed domain randomization, while a joint-temporal motion prior learns motion structure from mocap data.
A dual-mode mapping prior uses codebook-based classification for constrained
zero-shot reconstruction and continuous regression for flexible adaptation.
In classification, the frozen tokenizer regularizes adaptation with target codes; in regression, the motion prior optionally refines predictions for temporal consistency.

Our contributions are summarized as follows:
\begin{itemize}
    \item We introduce \textbf{mmSimPrior}, a simulation-pretrained
    framework that shifts most paired supervision from scarce real
    recordings to large-scale simulated radar--motion data, supporting
    zero-shot transfer and adaptation using only 24 paired real sequences.

    \item We develop a unified architecture that factorizes transferable
    knowledge into multi-modal signal, joint-temporal motion, and dual-mode
    mapping priors. Physics-informed domain randomization approximates
    propagation and acquisition shifts in heatmap space while preserving
    radar semantics; classification and regression support constrained
    zero-shot reconstruction and flexible real-data adaptation, respectively.

    \item We further construct a 4.2M-frame, 31K-sequence dataset suite
    and introduce a No-Overlap Setting that excludes repeated complete
    subject--environment--location--motion configurations across adaptation
    and test.
\end{itemize}

% \noindent
% Code and data will be released upon acceptance.
\section{Related Work}
\label{sec:related}

\subsection{Radar-Based Human Motion Reconstruction and Generalization}

Radar-based pose and mesh reconstruction has been studied using sparse
point clouds, Range--Doppler or Range--Angle heatmaps, multi-view heatmaps, and
high-dimensional radar tensors
~\cite{xue2021mmmesh,xue2022m4esh,lee2023hupr,
rahman2024mmvr,ho2024rt,kato2025raptr,fan2026m4human}.
Prior methods employ direct mesh regression, diffusion, weak supervision,
and radar-tensor modeling
~\cite{xue2021mmmesh,fan2024diffusion,kato2025raptr,ho2024rt}.
Recent datasets and methods have also considered cross-subject,
cross-environment, or unseen-activity evaluation
~\cite{xue2023towards,rahman2024mmvr,yang2023mm}.
Despite this progress, fine-grained radar-based human reconstruction
still relies heavily on paired real recordings collected under specific
subjects, environments, and sensing configurations.
Performance therefore remains closely tied to the available real-data
coverage and evaluation protocol.
mmSimPrior instead studies simulation-pretrained zero-shot and
limited-data transfer to real deployments, evaluated under a strict
no-overlap protocol.

\subsection{Simulation-Based Radar Supervision and Sim-to-Real Transfer}

Prior work reduces radar-data collection costs through physics-based
simulation, motion-capture-driven synthesis, video-conditioned
generation, and generative models
~\cite{xue2023towards,chen2023rf,deng2023midas,jiang2026podm,
huang2025one}.
Synthetic observations have been used to augment real-world training, support RF sensing, and enable radar-based activity or language
understanding
~\cite{xue2023towards,chen2023rf,lai2026radarllm,
yan2025mmexpert}.
mmSimPrior instead treats simulation as the primary source of paired
supervision rather than as supplementary augmentation.
Unlike prior radar-based human motion reconstruction methods trained mainly on paired
recordings from the deployment domain, we study whether simulation
can replace most such supervision for fine-grained motion
reconstruction, with either no data from the evaluation dataset or only 24 paired real sequences for adaptation.
To enable this shift, we factorize transferable knowledge into signal,
motion, and mapping priors and introduce physics-informed domain
randomization for robust real-world transfer.
\section{Radar Observation and Dataset Construction}
\label{sec:data}

This section first introduces the radar representations used throughout
the paper and then describes the construction of the large-scale simulated corpus used for pretraining.
Unlike RGB images, radar heatmaps are physically indexed response maps
derived from complex radio-frequency measurements.
To retain both radial kinematics and metric spatial structure, we
represent each radar frame using an RD heatmap together with three
orthogonal MVDR projections.

\subsection{Factorized Radar Heatmap Representation}
\label{sec:radar_representation}

\paragraph{FMCW radar measurements.}
At frame $t$, an FMCW radar records a complex IF tensor
$\mathbf X_t\in
\mathbb C^{N_{\rm tx}\times N_{\rm rx}\times N_c\times N_s}$,
indexed by transmitters, receivers, chirps, and ADC samples.
The IF tensor contains propagation-delay, radial-motion, and
inter-antenna phase information, but does not provide explicit
correspondences between measurements and human body parts.

\paragraph{Range--Doppler heatmap.}

By applying Range-FFT along the fast-time dimension and a
Doppler-FFT along the chirp dimension, the resulting Range--Doppler (RD) heatmap
$H_t^{\mathrm{RD}}$ represents reflected response magnitude as a
function of propagation distance and radial velocity.
Therefore, unlike an RGB pixel indexed by image-plane position, each RD
cell corresponds to a physically defined range--velocity bin.
The RD representation preserves radial kinematic evidence that is useful
for separating moving body responses from static or slowly varying
background structures.

\begin{figure}[t]
  \centering
  \includegraphics[width=0.65\linewidth]{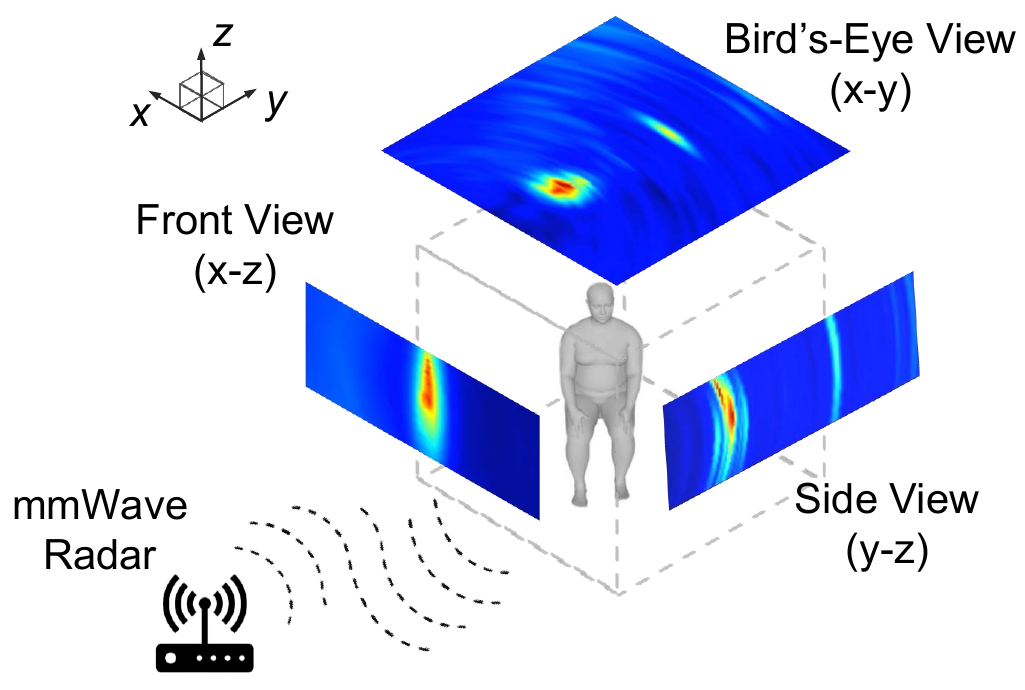}
  \caption{
\textbf{Illustration of multi-view MVDR spatial representation.}
We compute a 3D MVDR spatial heatmap around the human body and project it
into three orthogonal 2D views.
}
  \label{fig:mvdr}
\end{figure}

\paragraph{Multi-view MVDR spatial heatmaps.}

The RD heatmap does not directly resolve the 3D direction of a
reflection.
We therefore use phase differences across the virtual antenna array to
estimate spatial response power through Minimum Variance Distortionless
Response (MVDR) beamforming.
For a candidate location $\mathbf{p}=(x,y,z)$ on a discretized 3D grid,
the MVDR spectrum is
\begin{equation}
    P_t(\mathbf{p})
    =
    \left|
    \frac{1}{
        \mathbf{a}(\mathbf{p})^{H}
        \mathbf{R}_{t,r(\mathbf{p})}^{\dagger}
        \mathbf{a}(\mathbf{p})
    }
    \right|,
    \label{eq:mvdr}
\end{equation}
where $\mathbf{a}(\mathbf{p})$ is the steering vector associated with
the candidate position,
$\mathbf{R}_{t,r(\mathbf{p})}$ is the antenna-domain covariance matrix
at the corresponding range bin, and ${\dagger}$ denotes the Hermitian
pseudoinverse.
Intuitively, MVDR assigns high response power to locations whose expected
inter-antenna phase pattern agrees with the measured signal while
suppressing inconsistent interference.

Direct learning over the full 3D response volume is computationally
expensive and susceptible to sparse volumetric artifacts.
We instead factorize it into three orthogonal mean projections as shown in Fig.~\ref{fig:mvdr}. The front view captures horizontal--vertical body structure, the
bird's-eye view preserves ground-plane position and orientation, and the
side view provides complementary depth--height evidence.
Together, the factorized observation at frame $t$ is $ \mathbf{o}_t =
    \left\{
        H_t^{\mathrm{RD}},
        H_t^{\mathrm{front}},
        H_t^{\mathrm{bird}},
        H_t^{\mathrm{side}}
\right\}.$ Given a radar heatmap sequence $\mathbf{o}_{1:T}$, the reconstruction task is to
estimate the SMPL-X motion parameters $\Theta_{1:T}
=
\left\{
    \boldsymbol{\theta}_{1:T},
    \mathbf{t}_{1:T},
    \boldsymbol{\beta}
\right\},$
where $\boldsymbol{\theta}_t$ contains continuous 6D body-joint
rotations, $\mathbf{t}_t$ is the global root translation, and
$\boldsymbol{\beta}$ denotes sequence-level body shape.

\subsection{mmSimPrior-Sim: Physics-Based Radar--Motion Synthesis}
\label{sec:sim_dataset}

We construct mmSimPrior-Sim by driving a physics-based FMCW radar
simulator \cite{xue2023towards} with AMASS \cite{AMASS:ICCV:2019} motion sequences.
Each motion sequence is represented by a temporally varying SMPL-X mesh
and resampled to the radar frame rate.
Mesh triangles are treated as surface reflectors characterized by their
barycenters, surface normals, and area-dependent response strengths.

For a mesh triangle $q$, transmit antenna $i$, receive antenna $j$, and
frequency sample $n$, its bistatic reflection can be expressed as
\begin{equation}
    s_{qijn}
    =
    \frac{A_q G_{qij}}
    {d^{\mathrm{tx}}_{qi}d^{\mathrm{rx}}_{qj}}
    \exp
    \left(
        j2\pi(f_c+\Delta f_n)\tau_{qij}
    \right),
    \label{eq:simulated_reflection}
\end{equation}
where $A_q$ represents the surface-area contribution,
$d^{\mathrm{tx}}_{qi}$ and $d^{\mathrm{rx}}_{qj}$ are the
transmit- and receive-side path lengths,
$\tau_{qij}$ is the corresponding propagation delay, and $G_{qij}$
models the dependence of reflection strength on surface orientation.
Geometrically invalid back-facing responses are removed, and the
remaining triangle responses are accumulated to synthesize the complex
IF signal.

Each motion is rendered at multiple radar--subject distances to introduce
distance-dependent variation in attenuation, angular resolution, and
body-reflection patterns.
This yields synchronized radar heatmaps paired with SMPL-X motion annotations.
The resulting radar--motion pairs provide scalable supervision for
learning the signal and mapping priors, while their underlying AMASS
motions are used to pretrain the tokenized motion prior.
\section{Methodology}
\label{sec:method}

\begin{figure}[ht]
  \centering
  \includegraphics[width=0.98\linewidth]{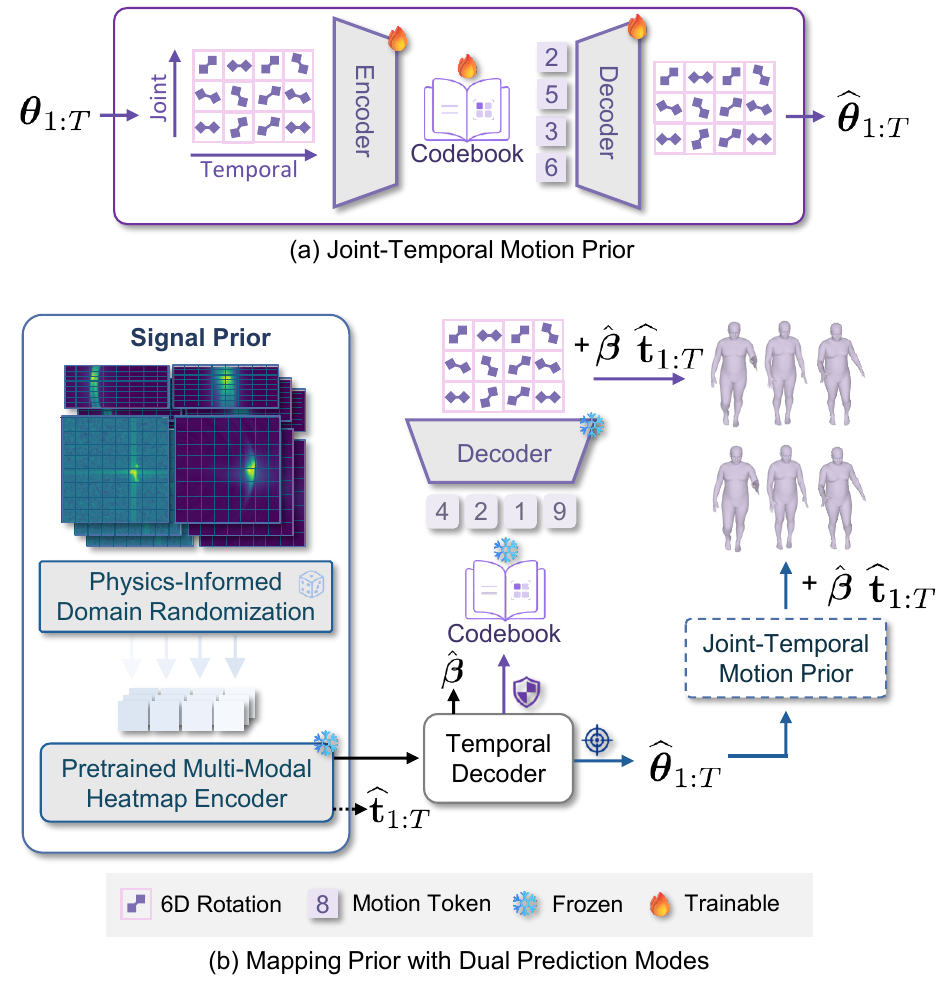}
    \caption{
\textbf{Overview of mmSimPrior.}
(a) Joint-temporal tokenization learns the motion prior.
(b) Domain-randomized radar heatmaps are encoded by the signal prior
and decoded through classification- or regression-based mapping modes.
}
  \label{fig:overview}
\end{figure}

The goal of mmSimPrior is to recover SMPL-X pose, translation, and shape
from factorized radar heatmap sequences.
As shown in Fig.~\ref{fig:overview}, a signal prior encodes synchronized
RD and multi-view MVDR heatmaps, a joint-temporal motion prior models
human motion structure, and a shared mapping prior connects the two.
The mapping prior supports codebook-based classification for constrained
zero-shot reconstruction and continuous regression for flexible
real-world adaptation.

\subsection{Physics-Informed Domain Randomization}
\label{sec:domain_randomization}

\begin{figure}[ht]
  \centering
  \includegraphics[width=\linewidth]{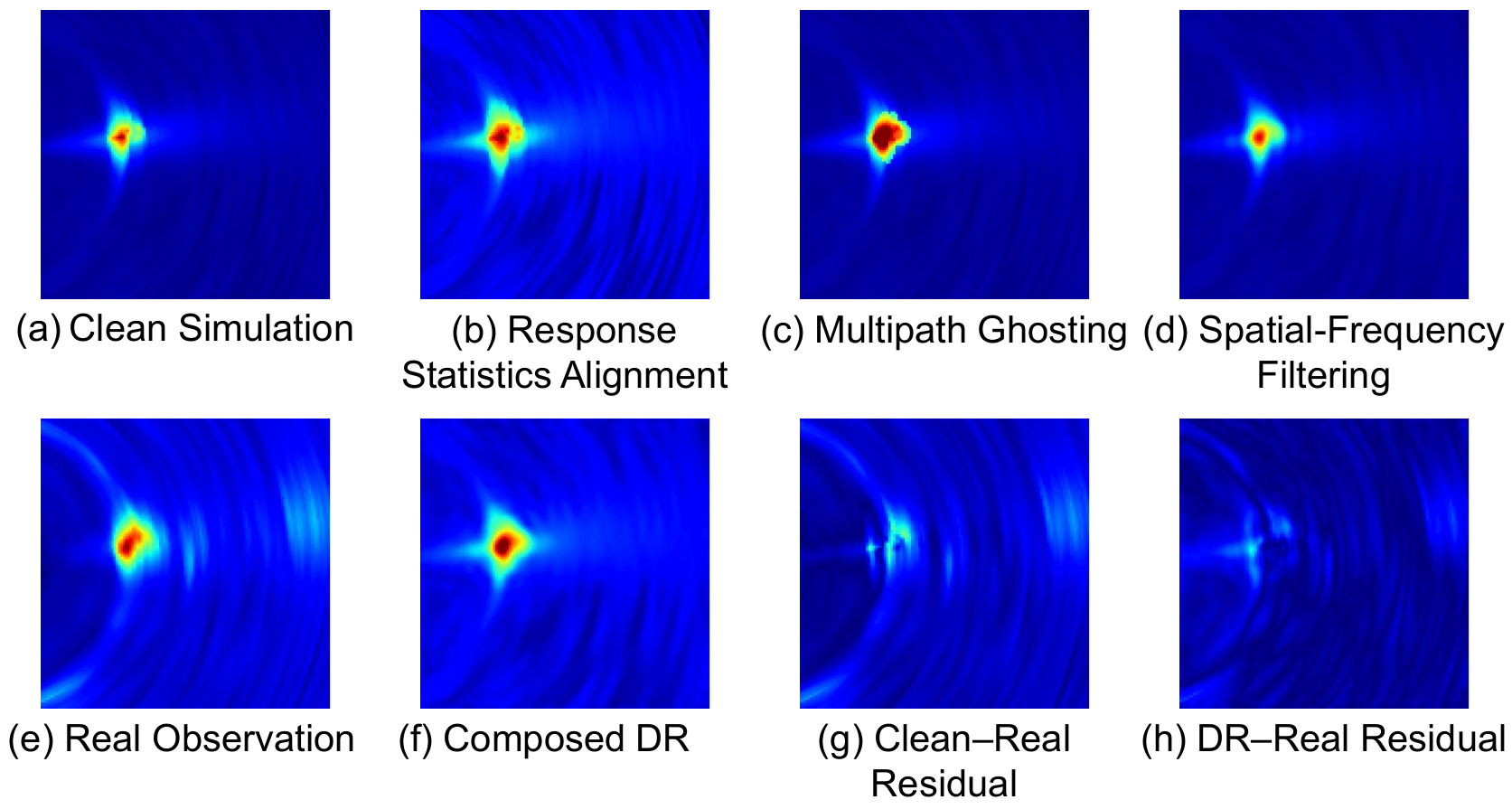}
  \caption{
    \textbf{Physics-informed domain randomization.}
    Modality-conditioned perturbations preserve the dominant human response
    while reducing the discrepancy between matched simulated and real MVDR
    heatmaps.
    }
  \label{fig:domain_randomization_vis}
\end{figure}

Although simulated and real IF signals share the same processing
pipeline, their heatmaps can still differ because a clean simulator
cannot capture the full variability of multipath, clutter, and
acquisition statistics. Exhaustively resimulating diverse propagation
and acquisition conditions is costly, while generic image
transformations cannot be applied indiscriminately: each radar
modality has axes with distinct range, Doppler, or metric spatial
meanings. We therefore introduce physics-informed domain
randomization in radar heatmap space. Rather than repeatedly
simulating the complete propagation process, we construct efficient,
modality-conditioned surrogates whose forms, directions, and modality
assignments are tied to specific propagation and acquisition phenomena.
As illustrated in Fig.~\ref{fig:domain_randomization_vis}, these transformations diversify low-level
radar appearance while preserving the dominant human response and its
paired motion annotation.

For a simulated heatmap sequence $\mathbf{o}_{1:T}$, we generate
\begin{equation}
    \widetilde{\mathbf{o}}_{1:T}
    =
    \mathcal{A}_{\boldsymbol{\xi}}
    \left(
        \mathbf{o}_{1:T}
    \right),
    \qquad
    \boldsymbol{\xi}\sim q_s(\boldsymbol{\xi}),
    \label{eq:domain_randomization}
\end{equation}
where $\mathcal{A}_{\boldsymbol{\xi}}$ denotes a
modality-conditioned composition of label-preserving transformations,
and $q_s$ controls their occurrence and severity at training step $s$.
These label-preserving perturbations expose the model to
propagation- and acquisition-level variations before deployment.

\paragraph{Multipath ghosting.}
Indirect propagation paths can generate displaced and attenuated replicas
of strong human responses.
We emulate this phenomenon by shifting high-energy regions along
view-dependent spatial axes:
\begin{equation}
    \tau_{\mathrm{ghost}}(H)
    =
    H
    +
    \alpha(s)
    \mathcal{S}_{\Delta}
    \left(
        H\odot\mathbb{I}[H>\rho]
    \right),
    \label{eq:ghosting}
\end{equation}
where $\mathcal{S}_{\Delta}$ applies a spatial shift,
$\rho$ selects dominant responses, and $\alpha(s)$ controls the replica
strength.
The shift direction is conditioned on the physical axes represented by
each MVDR view.

\paragraph{Resolution degradation.}
Finite angular resolution and increasing sensing distance can spread a
localized reflection over neighboring spatial cells.
We approximate this effect by applying scheduled Gaussian spreading to spatial responses$\tau_{\mathrm{spread}}(H)=H +\beta(s)
\left(
    \mathbb{I}[H>\rho]*G_{\sigma}
\right)$, where $G_{\sigma}$ is a Gaussian kernel.
This transformation reduces reliance on simulation-specific sharp
spatial responses.

\paragraph{Acquisition-level variations.}
We additionally perturb spatial-frequency content using
$\tau_{\mathrm{freq}}(H)
=\mathcal{F}^{-1}(M_{\omega_c}\odot\mathcal{F}(H))$
and apply monotone intensity remapping
$\tau_{\mathrm{stat}}(H)=g_{\eta}(H)$
to capture acquisition-dependent variations in antenna response, gain,
clutter, and noise-floor statistics.
Both transformations alter low-level radar appearance while preserving
the dominant spatial response structure.

\paragraph{Progressive nuisance curriculum.}
Applying severe randomization from the beginning can obscure the
radar--motion correspondence.
We therefore use a piecewise curriculum,
$\boldsymbol{\phi}(s)=\boldsymbol{\phi}_i$ for
$s\in[s_i,s_{i+1})$, to progressively increase selected nuisance
strengths, including multipath replicas and peak spreading.
This allows the model to first learn stable correspondences before being
exposed to stronger deployment variations.

\subsection{Multi-Modal Radar Signal Prior}
\label{sec:signal_prior}

The four radar modalities provide complementary cues but differ in
resolution and aspect ratio.
For modality $k$, a modality-specific adapter produces an equal token
budget,
$\mathbf{U}^{(k)}_t=f^{(k)}_{\mathrm{patch}}(H^{(k)}_t)
\in\mathbb{R}^{N\times d}$.
The padded front and side views use anisotropic patches, whereas RD and
bird's-eye use square patches.

We pretrain the adapters and a shared Transformer as a multi-modal masked
autoencoder on domain-randomized simulation.
We independently mask each modality and jointly encode all visible
tokens with a shared Transformer, allowing attention to associate RD
kinematics with MVDR spatial responses.
Modality-specific reconstruction heads are used only during pretraining;
the shared encoder is retained as the signal prior.

\subsection{Joint-Temporal Motion Prior}
\label{sec:motion_prior}

To constrain this ambiguous radar-based motion reconstruction, we learn a motion prior from SMPL-X motion sequences using a VQ-VAE~\cite{van2017neural}.
A motion sequence is represented by continuous 6D rotations $\mathbf{R}_{1:T} \in \mathbb{R}^{T\times J\times6},$
including the root orientation and body-joint rotations.
The motion encoder maps the rotation sequence to a joint-temporal latent
lattice $\mathbf{E}=E_{\mathrm{mot}}\left(    \mathbf{R}_{1:T}\right)\in\mathbb{R}^{T'\times J'\times d_c}$, where $T'$ and $J'$ denote the reduced temporal and joint-axis
resolutions, respectively.
Each latent vector is assigned to its nearest entry in a learned
codebook
$\mathcal{C}=\{\mathbf{c}_m\}_{m=1}^{|\mathcal{C}|}$:
\begin{equation}
    z_{t,j}
    =
    \arg\min_m
    \left\|
        \mathbf{E}_{t,j}
        -
        \mathbf{c}_m
    \right\|_2^2.
    \label{eq:motion_quantization}
\end{equation}
A motion decoder reconstructs the continuous rotation sequence from the
quantized joint-temporal representation $\widehat{\mathbf{R}}_{1:T}
    =
    D_{\mathrm{mot}}
    \left(
        \{
            \mathbf{c}_{z_{t,j}}
        \}
    \right).$
The tokenizer is trained with rotation-reconstruction, temporal,
joint-geometry, and commitment losses, after which its codebook and
decoder are frozen.

The motion prior serves asymmetric roles across the two modes of mapping prior.
For Classification mode (Cls), its codebook and decoder form the primary prediction path, while
its frozen tokenizer additionally provides target code indices during
real-world adaptation.
For Regression mode (Reg), the prior is used only as an optional refinement for temporal
consistency.

\subsection{Dual-Space Radar-Conditioned Motion Decoding}
\label{sec:dual_space_decoding}

The mapping prior first temporally aligns frame-level heatmap features
$\mathbf{U}_{1:T}$ as
$\widetilde{\mathbf{U}}_{1:T'}
=g_{\mathrm{align}}(\mathbf{U}_{1:T})$,
and aggregates temporal context as
$\mathbf{F}_{1:T'}=g_{\mathrm{temp}}
(\widetilde{\mathbf{U}}_{1:T'})$.
At each reduced time step, a shared temporal query cross-attends to the
multi-modal context,
$\mathbf{m}_t=g_{\mathrm{query}}(\mathbf{q},\mathbf{F}_t)$.
Both modes share this decoder and differ only in their output space.

\paragraph{Classification-based mode.}
The Cls head predicts codebook logits
$\boldsymbol{\ell}_{t,j}=h_{\mathrm{Cls}}(\mathbf{m}_t)_j$.
Following previous work~\cite{dwivedi2024tokenhmr,saleem2025genhmr},
we obtain code probabilities
$\boldsymbol{\pi}_{t,j}
=\operatorname{softmax}(\boldsymbol{\ell}_{t,j})$
and perform differentiable soft code retrieval:
\begin{equation}
\begin{aligned}
    \widetilde{\mathbf{c}}_{t,j}
    &=
    \sum_{m=1}^{|\mathcal{C}|}
    \boldsymbol{\pi}_{t,j}[m]\mathbf{c}_m,\\
    \widehat{\mathbf{R}}^{\mathrm{Cls}}_{1:T}
    &=
    D_{\mathrm{mot}}
    \left(
        \{\widetilde{\mathbf{c}}_{t,j}\}_{t,j}
    \right).
\end{aligned}
\label{eq:classification_decoding}
\end{equation}

The frozen codebook and decoder constrain predictions toward learned
joint-temporal motion patterns.

\paragraph{Regression-based mode.}
The Reg head directly predicts continuous body rotations,
$\widehat{\mathbf R}^{\mathrm{Reg}}_{1:T}
=h_{\mathrm{Reg}}(\mathbf m_{1:T'})$,
retaining greater flexibility for real-domain correction.
Optionally, the Reg output can be projected through the frozen motion
prior to improve temporal smoothness.

Both modes are trained using SMPL-X supervision:
\begin{equation}
\begin{aligned}
    \mathcal{L}_{\mathrm{map}}
    ={}&
    \lambda_{\theta}\mathcal{L}_{\theta}
    +
    \lambda_{\mathrm{joint}}\mathcal{L}_{\mathrm{joint}}
    +
    \lambda_{\mathrm{trans}}\mathcal{L}_{\mathrm{trans}}
    +
    \lambda_{\beta}\mathcal{L}_{\beta}\\
    &+
    \lambda_{\mathrm{dyn}}\mathcal{L}_{\mathrm{dyn}},
\end{aligned}
\label{eq:mapping_objective}
\end{equation}
where the losses supervise body rotations, root-relative joints,
global translation, body shape, and translation dynamics.

\subsection{Motion-Prior-Guided Real-World Adaptation}
\label{sec:real_adaptation}

Given a small paired real set $\mathcal{D}_{\mathrm{ft}}$, we freeze the
pretrained signal-encoder weights, insert LoRA adapters into its
Transformer blocks~\cite{hu2021lora}, and jointly optimize the adapters
with the temporal mapping modules and prediction heads.

For Cls, the frozen tokenizer converts ground-truth motion into target
indices $z^{*}_{t,j}$, and we optimize
$\mathcal{L}^{\mathrm{Cls}}_{\mathrm{ft}}
=\mathcal{L}_{\mathrm{map}}
+\lambda_{\mathrm{tok}}\mathcal{L}_{\mathrm{tok}}$, where
$\mathcal{L}_{\mathrm{tok}}
=\frac{1}{T'J'}\sum_{t,j}
\operatorname{CE}(\boldsymbol{\ell}_{t,j},z^{*}_{t,j})$.
We linearly warm $\lambda_{\mathrm{tok}}$  and
refer to this token supervision as motion-prior-guided (MPG)
adaptation.
Reg follows the same strategy without $\mathcal{L}_{\mathrm{tok}}$.
\section{Experiments}
\label{sec:experiments}

% ---------------------------------------------------------------
\subsection{Experimental Setup}
\label{subsec:exp_setup}

\noindent \textbf{Benchmarks and inputs.}
We use \textbf{mmSimPrior-Sim} as the AMASS-driven simulated radar--motion corpus for pretraining, and evaluate on two real-world benchmarks. 
\textbf{mmSimPrior-Real} denotes our collected real-world benchmark with three indoor environments. 
We further evaluate on the public \textbf{RT-Pose} benchmark~\cite{ho2024rt} as an external validation dataset. Because RT-Pose provides only 3D keypoints, we derive pseudo-SMPL-X annotations by fitting RGB-based human mesh recovery
estimates~\cite{wang2025prompthmr,yang2026sam} to the provided keypoints.
Unless otherwise specified, all methods are pretrained on mmSimPrior-Sim and evaluated on real radar heatmaps under zero-shot or limited-data finetuning protocols.

\noindent \textbf{mmSimPrior-Real annotations.}
mmSimPrior-Real contains recordings from three indoor environments
(Hall, Lab, and Studio), covering six volunteers, 40 motion categories,
and three sensing-distance ranges; the protocol was approved by our
Institutional Review Board, and all participants provided informed consent.

\noindent \textbf{No-overlap protocol and metrics.}
On mmSimPrior-Real, we evaluate using a No-Overlap Setting (NOS): no finetuning-test sequence pair shares the same complete setting across subject identity, environment, sensing location, and motion category. 
Unless otherwise specified, limited finetuning uses 24 real sequences; Zero-shot denotes evaluation without any annotation, paired supervision, or model selection from the evaluated benchmark.  
We report MPJPE and PA-MPJPE, W-MPJPE when evaluating world-space localization, and MPVPE for mesh-level evaluation.
Avg. denotes a sequence-weighted average.

\noindent \textbf{Baselines.}
We compare against adapted radar-based human motion reconstruction baselines, including mmDiff$^\dagger$\cite{fan2024diffusion}, RAPTR$^\dagger$~\cite{kato2025raptr}, and mmGPE$^\dagger$~\cite{xue2023towards}. 
For fairness, all adapted baselines use the same four-modality input, simulated pretraining corpus, finetuning budget, evaluation splits, and metrics as mmSimPrior. 
Baseline adaptation details, metric definitions, averaging formula, and split construction are provided in the supplementary material.

\begin{table*}[t]
\centering
\small
\setlength{\tabcolsep}{2.6pt}
\renewcommand{\arraystretch}{1.12}

\begin{tabular*}{\textwidth}{
@{\extracolsep{\fill}} l|ccc|ccc|ccc @{}
}
\toprule
\multirow{2}{*}{Method}
& \multicolumn{3}{c|}{Hall}
& \multicolumn{3}{c|}{Lab}
& \multicolumn{3}{c}{Studio} \\
& MPJPE$\downarrow$ & PA-MPJPE$\downarrow$ & W-MPJPE$\downarrow$
& MPJPE$\downarrow$ & PA-MPJPE$\downarrow$ & W-MPJPE$\downarrow$
& MPJPE$\downarrow$ & PA-MPJPE$\downarrow$ & W-MPJPE$\downarrow$ \\
\midrule
mmDiff$^\dagger$
& 104.0 & 74.7 & 173.4
& 102.5 & 78.6 & 160.2
& 104.8 & 80.4 & 167.4 \\
RAPTR$^\dagger$
& 115.9 & 87.1 & 174.8
& 107.8 & 85.1 & 171.0
& 118.3 & 93.3 & 172.9 \\
mmGPE$^\dagger$
& 106.4 & 75.5 & 147.0
& 104.7 & 76.0 & 144.9
& 101.9 & 74.7 & 135.6 \\
\midrule
\textbf{mmSimPrior-Reg}
& \textbf{75.9} & \textbf{54.7} & \textbf{116.0}
& \textbf{77.2} & \textbf{57.0} & \textbf{116.0}
& \textbf{62.2} & \textbf{49.5} & \textbf{95.0} \\
\textbf{mmSimPrior-Cls}
& \underline{89.5} & \underline{65.5} & \underline{125.6}
& \underline{92.4} & \underline{68.5} & \underline{127.5}
& \underline{86.2} & \underline{65.8} & \underline{114.2} \\
\bottomrule
\end{tabular*}

\caption{
\textbf{Same-environment 24-sequence NOS finetuning on mmSimPrior-Real.}
All methods use the same four-modality radar heatmap input, mmSimPrior-Sim pretraining corpus, finetuning budget, and NOS split. The best / second best results are in \textbf{boldface}, and \underline{underlined}, respectively.
Metrics are in millimeters; lower is better.
$^\dagger$ denotes adapted baselines pretrained on mmSimPrior-Sim.
}
\label{tab:same_env_main}
\end{table*}

% ---------------------------------------------------------------
\subsection{Overall Result}
\label{subsec:overall_result}

\noindent \textbf{Performance on mmSimPrior-Real.}
Table~\ref{tab:same_env_main} evaluates 24-sequence NOS finetuning
within each environment.
Although adaptation and testing occur in the same environment, the support set contains only 24 paired sequences and shares no exact subject–location–motion configuration with the test set.
mmSimPrior-Reg achieves the best result across all three metrics and
environments, while mmSimPrior-Cls consistently ranks second.
Relative to the strongest adapted MPJPE baseline, Reg reduces error
by 27.0\%, 24.7\%, and 39.0\% in Hall, Lab, and Studio,
respectively.
The consistent improvements in MPJPE, PA-MPJPE, and W-MPJPE show
that the learned priors improve both articulated pose recovery and
world-space localization under limited-data adaptation.
The stronger Reg results further indicate that continuous decoding
can effectively absorb target-domain corrections from limited real
supervision. These results support the central premise that simulation can provide
most paired supervision, with only limited real data needed for
deployment-specific correction.

\begin{figure}[t]
  \centering
  \includegraphics[width=0.47\textwidth]{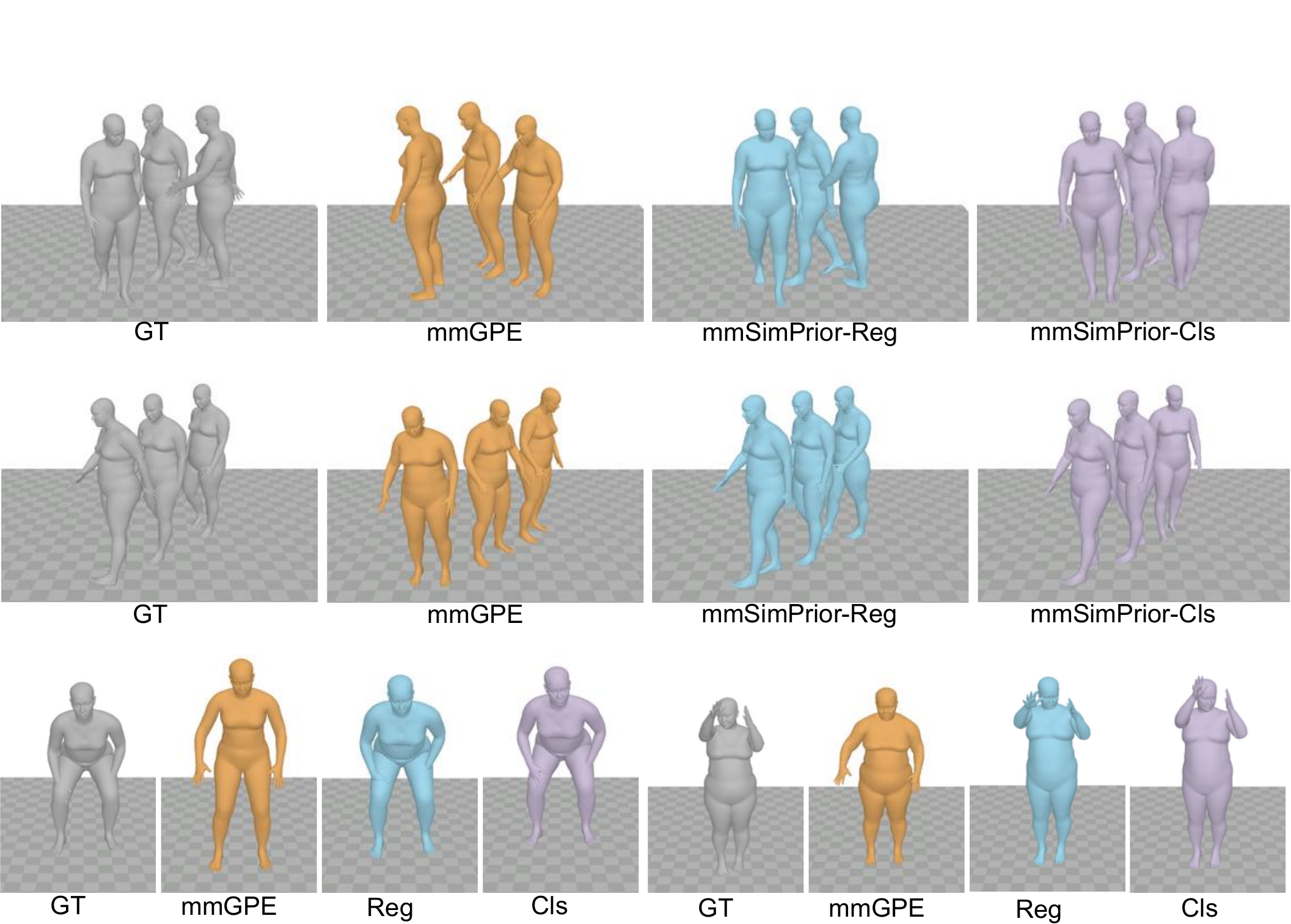}
  \caption{
  \textbf{Qualitative comparison on real-world radar sequences.}
  We compare
  \protect\textcolor{mmgpeText}{mmGPE},
  \protect\textcolor{regText}{mmSimPrior-Reg}, and
  \protect\textcolor{clsText}{mmSimPrior-Cls}
  with the
  \protect\textcolor{gtText}{SMPL-X ground truth}.
}
  \label{fig:qualitative_results}
\end{figure}

Figure~\ref{fig:qualitative_results} shows that both mmSimPrior
variants recover motion progression and local articulation more
faithfully than mmGPE.
Qualitatively, they also maintain more stable foot--ground contact
with substantially less foot sliding.
Beyond frame-level reconstruction accuracy, the optional motion-prior
refinement acts as a targeted temporal regularizer for Reg. It reduces joint-velocity, acceleration, and jerk errors by 7.7\%, 18.0\%, and 36.7\%, respectively, while changing all spatial reconstruction metrics by at most 1.3\% (see the supplementary material).

% RT-pose----------------------------------------------------------------------------------------------
\begin{table}[t]
\centering
\small
\renewcommand{\arraystretch}{1.15}

\begin{tabular*}{\columnwidth}{
@{\extracolsep{\fill}} l|ccc @{}
}
\toprule
Method
& MPJPE$\downarrow$
& PA-MPJPE$\downarrow$
& MPVPE$\downarrow$ \\
\midrule
mmDiff$^\dagger$ & 126.0 & 81.6 & 165.8 \\
RAPTR$^\dagger$  & 131.0 & 89.3 & 171.0 \\
mmGPE$^\dagger$  & 126.1 & 82.5 & 173.0 \\
\midrule
\textbf{mmSimPrior-Reg}
& \textbf{116.9}
& \textbf{78.1}
& \textbf{153.2} \\
\textbf{mmSimPrior-Cls}
& \underline{119.9}
& \underline{80.8}
& \underline{156.2} \\
\bottomrule
\end{tabular*}

\caption{\textbf{Minute-level external validation on
RT-Pose~\cite{ho2024rt}.} The best / second best results are in \textbf{boldface}, and \underline{underlined}, respectively. 
Metrics are in millimeters; lower is better.}
\label{tab:rtpose_public}
\end{table}

\noindent \textbf{External validation on RT-Pose.}
Table~\ref{tab:rtpose_public} uses the same minute-level support budget
for all methods. mmSimPrior-Reg achieves the best MPJPE, PA-MPJPE, and
MPVPE, improving over the strongest baseline by 9.1, 3.5, and
12.6\,mm, respectively, while mmSimPrior-Cls ranks second. Because
RT-Pose uses a different radar platform, these gains also support
cross-device transfer under limited adaptation. 

\begin{figure}[ht]
  \centering
  \includegraphics[width=0.47\textwidth]{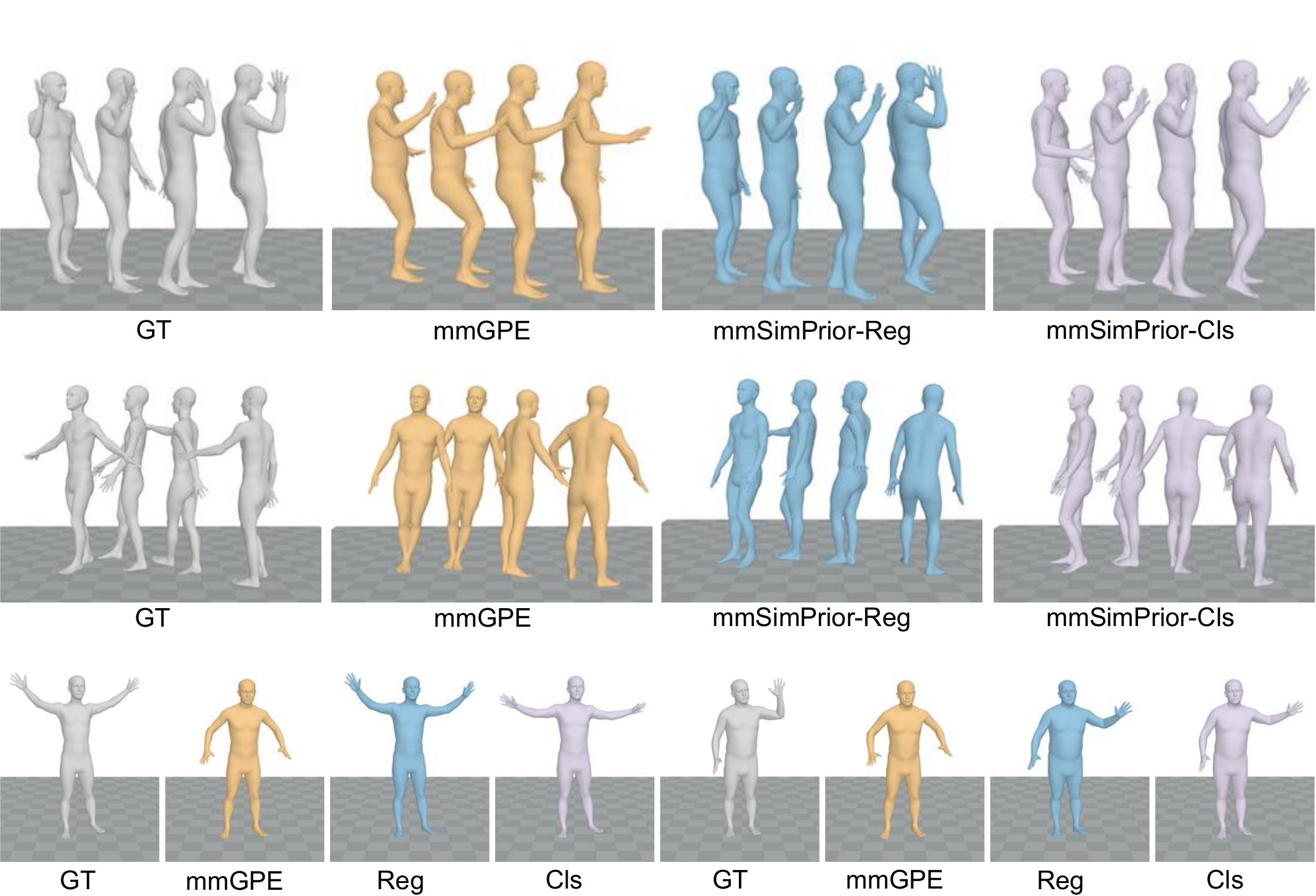}
  \caption{
  \textbf{Qualitative comparison on RT-Pose.}
  We compare
  \protect\textcolor{mmgpeText}{mmGPE},
  \protect\textcolor{regText}{mmSimPrior-Reg}, and
  \protect\textcolor{clsText}{mmSimPrior-Cls}
  with
  \protect\textcolor{gtText}{pseudo-SMPL-X Ground Truth}.
}
  \label{fig:qualitative_rtpose}
\end{figure}

Figure~\ref{fig:qualitative_rtpose} shows that these advantages persist
under the RT-Pose domain shift.
Both mmSimPrior variants better preserve the temporal action phase
and maintain more stable foot placement, resulting in visibly less
foot sliding.
Reg more closely matches fine-grained limb configurations, while Cls
avoids implausible intermediate poses by decoding through the learned
motion codebook.
Together with Table~\ref{tab:rtpose_public}, these results show that
the learned priors improve both reconstruction accuracy and temporal
plausibility.

% ---------------------------------------------------------------
\subsection{Generalization Across Unseen Environments}
\label{subsec:env_gen}

We next study robustness to environment shift by evaluating transfer across previously unseen real-world environments.
Compared with same-environment evaluation, this setting is more challenging because the test environment can introduce unseen clutter, multipath patterns, and reflective structures.

\begin{table}[t]
\centering
\small
\setlength{\tabcolsep}{2.5pt}
\renewcommand{\arraystretch}{1.3}

\begin{tabular*}{\columnwidth}{
@{\extracolsep{\fill}} l|cc|cc @{}
}
\toprule
\multirow{2}{*}{Method}
& \multicolumn{2}{c|}{Zero-shot Avg.}
& \multicolumn{2}{c}{24-seq finetune} \\
& MPJPE & PA-MPJPE
& MPJPE & PA-MPJPE \\
\midrule
mmDiff$^\dagger$
& 130.0 & 97.2 & 117.1 & 87.1 \\
RAPTR$^\dagger$
& 170.2 & 104.0 & 119.0 & 90.4 \\
mmGPE$^\dagger$
& 141.3 & 97.0 & 111.8 & 81.0 \\
\midrule
\textbf{mmSimPrior-Reg}
& \underline{119.1} & \underline{93.1}
& \textbf{96.7} & \textbf{73.3} \\
\textbf{mmSimPrior-Cls}
& \textbf{118.9} & \textbf{89.7}
& \underline{100.0} & \underline{75.2} \\
\bottomrule
\end{tabular*}

\caption{
\textbf{Unseen-environment transfer on mmSimPrior-Real.}
Zero-shot reports sequence-weighted averages over all splits without finetuning. The 24-seq columns report Hall$\rightarrow$Lab cross-environment finetuning, where models are finetuned on Hall and evaluated on the no-overlap Lab split.
The best / second best results are in \textbf{boldface}, and \underline{underlined},
respectively. Metrics are in millimeters; lower is better.
}
\label{tab:cross_env_main}
\end{table}

\noindent \textbf{Unseen-environment performance.}
Table~\ref{tab:cross_env_main} evaluates two deployment regimes:
direct zero-shot transfer and Hall$\rightarrow$Lab adaptation without
using any Lab sequence for finetuning.
In the zero-shot setting, mmSimPrior-Cls achieves the best results,
reducing MPJPE / PA-MPJPE over the strongest baseline by
8.5\% / 7.5\%.
Its frozen codebook and motion decoder constrain uncertain radar
features to plausible joint-temporal patterns when target-domain
statistics are unavailable.
After finetuning on 24 Hall sequences, mmSimPrior-Reg becomes the
strongest mode and reduces the two errors by 13.5\% / 9.5\%.
Continuous decoding is less constrained by the discrete motion space
and therefore better exploits limited real supervision for
domain-specific pose correction.
This mode switch matches their intended roles: Cls provides a stronger
structural constraint without target supervision, whereas Reg better
absorbs domain-specific corrections when limited real data are available.

\subsection{Ablation Studies}
\label{subsec:ablation}

% ---------------------------------------------------------------
\begin{table}[t]
\centering
\small
\setlength{\tabcolsep}{1.15pt}
\renewcommand{\arraystretch}{1.2}

\begin{tabular*}{\columnwidth}{
@{\extracolsep{\fill}} lccccccc @{}
}
\toprule
& & & &
\multicolumn{2}{c}{Zero-shot}
& \multicolumn{2}{c}{24-seq finetune} \\
\cmidrule(lr){5-6}
\cmidrule(lr){7-8}
Mode & Sim. & DR & MPG
& MPJPE & PA
& MPJPE & PA \\
\midrule
\multirow{3}{*}{Reg}
& $\times$ & $\times$ & --
& -- & -- & 130.2 & 90.9 \\
& $\checkmark$ & $\times$ & --
& 140.6 & 90.5 & 126.6 & 85.4 \\
& $\checkmark$ & $\checkmark$ & --
& \textbf{129.4} & \textbf{85.4}
& \textbf{75.9} & \textbf{54.7} \\
\midrule
\multirow{4}{*}{Cls}
& $\times$ & $\times$ & $\times$
& -- & -- & 120.8 & 88.7 \\
& $\checkmark$ & $\times$ & $\times$
& 159.5 & 113.1 & 101.7 & 77.1 \\
& $\checkmark$ & $\checkmark$ & $\times$
& \textbf{123.9} & \textbf{84.4}
& 94.9 & 66.6 \\
& $\checkmark$ & $\checkmark$ & $\checkmark$
& \textbf{123.9} & \textbf{84.4}
& \textbf{89.5} & \textbf{65.5} \\
\bottomrule
\end{tabular*}

\caption{
\textbf{Ablation on the Hall split of mmSimPrior-Real.}
Sim. denotes pretraining on mmSimPrior-Sim, DR denotes
physics-informed domain randomization during simulated
pretraining, and MPG denotes motion-prior-guided adaptation.
The best / second best results are in \textbf{boldface}, and \underline{underlined}, respectively. 
Metrics are in millimeters; lower is better.
}
\label{tab:ablation_main}
\end{table}

Table~\ref{tab:ablation_main} isolates the effects of simulation
pretraining, physics-informed domain randomization (DR), and
Motion-Prior-Guided (MPG) adaptation.
Simulation pretraining and DR are complementary: the former supplies
broad paired motion coverage, whereas the latter prevents the learned
signal and mapping priors from overfitting simulation-specific
response statistics.
Clean simulation pretraining improves finetuned MPJPE / PA-MPJPE by
2.8\% / 6.1\% for Reg and 15.8\% / 13.1\% for Cls, while enabling
zero-shot inference.
Adding DR produces the dominant gains: 8.0\% / 5.6\% zero-shot and
40.0\% / 35.9\% finetuned improvements for Reg, and
22.3\% / 25.4\% zero-shot improvements for Cls.
Thus, synthetic scale alone is insufficient; DR makes the learned
simulation priors transferable.
Finally, MPG improves finetuned Cls by a further
5.7\% / 1.7\%, confirming the benefit of frozen-tokenizer
supervision.

\section{Conclusion}
\label{sec:conclusion}

We presented \textbf{mmSimPrior}, a sim-to-real framework for
data-efficient radar-based human motion reconstruction.
By shifting paired supervision from scarce real recordings to large-scale
simulated radar--motion data, mmSimPrior learns transferable signal,
motion, and mapping priors.
Physics-informed domain randomization improves their robustness to
real-world propagation and acquisition shifts, while the
classification- and regression-based mapping modes provide
complementary trade-offs between constrained zero-shot reconstruction
and flexible real-data adaptation.
The motion prior further guides limited-data Cls adaptation through
frozen-tokenizer supervision.
Experiments under the No-Overlap Setting, together with external
validation on RT-Pose, demonstrate consistent transfer across
same-environment, unseen-environment, and cross-dataset settings with no or limited supervision.
Future work will extend this framework to more complex human scenes and
broader real-world sensing conditions.

% Prevent main-paper floats from entering the references.
\FloatBarrier

% Keep references where they currently appear, so the main-paper
% layout changes as little as possible.
\bibliography{references}

% =========================================================
% Supplementary material
% =========================================================

% This page break is for arXiv only.
\clearpage

% Full-width supplementary title.
\makeatletter
\twocolumn[
\begin{@twocolumnfalse}
\begin{center}
\begin{minipage}{0.96\textwidth}
\centering

{\LARGE\bfseries
Supplementary Material for\par}

\vspace{0.35em}

{\LARGE\bfseries
mmSimPrior: Learning Simulation Priors for Data-Efficient
Real-World Generalizable Radar-Based Human Motion Reconstruction\par}

\end{minipage}
\end{center}

\vspace{1.2em}
\end{@twocolumnfalse}
]
\makeatother

\appendix

% Use S1, S2, ... for supplementary figures/tables/equations.
\setcounter{figure}{0}
\setcounter{table}{0}
\setcounter{equation}{0}

\renewcommand{\thefigure}{S\arabic{figure}}
\renewcommand{\thetable}{S\arabic{table}}
\renewcommand{\theequation}{S\arabic{equation}}

\section{Implementation Details}
\label{supp:implementation_details}

mmSimPrior operates on 32-frame clips at 10\,Hz and receives one Range--Doppler (RD) heatmap together with front, bird's-eye, and side MVDR heatmaps. Training comprises signal-prior pretraining, motion-prior pretraining, radar-conditioned mapping pretraining, and
limited-data real-world adaptation. 

\paragraph{Multi-modal signal prior.}
RD and bird's-eye inputs have resolution $128\times128$; front and side inputs have resolution $128\times50$ and are symmetrically padded to $128\times64$. Patch sizes are $16\times16$ for RD and bird's-eye views and $16\times8$ for front and side views, yielding 64 tokens per modality and 256 tokens per frame. The masked autoencoder uses a 12-layer shared
Transformer encoder (width 768, 12 heads) and four-layer
modality-specific decoders (width 512, eight heads). We independently mask 75\% of patches in each modality and optimize the mean masked-patch reconstruction loss for 200K iterations with AdamW, batch size 256, learning rate $2\times10^{-5}$, and weight decay $0.05$.

\paragraph{Joint-temporal motion prior.}
The motion tokenizer receives 32 frames of 6D rotations for 22 SMPL-X
body joints and encodes them into an $8\times40$ joint-temporal latent
lattice containing 320 motion tokens.
Each token is quantized using an EMA-updated codebook with 2,048
entries of dimension 256.
The encoder and decoder use width 512 and two residual blocks.
The tokenizer objective is
$20\mathcal{L}_{\mathrm{rot}}
+10\mathcal{L}_{\mathrm{vel}}
+100\mathcal{L}_{\mathrm{joint}}
+\mathcal{L}_{\mathrm{com}}$,
where the terms supervise 6D rotations, temporal differences,
SMPL-X joint geometry, and codebook commitment, respectively.
We train the tokenizer for 200K iterations using AdamW with batch size
256 and learning rate $2\times10^{-4}$.
The entire motion tokenizer is frozen in all subsequent stages.

\paragraph{Radar-conditioned mapping prior.}
Two stride-two temporal reductions compress the 32-frame signal features into eight temporal groups. A four-layer temporal Transformer and a shared six-layer cross-attention decoder aggregate the radar context.
Cls predicts 40 codebook distributions per temporal group (320 in total) and decodes their probability-weighted embeddings through the frozen motion decoder, whereas Reg directly predicts 32-frame 6D rotations. Both modes also predict framewise root translation and one sequence-level ten-dimensional shape vector. Mapping losses supervise rotation, root-relative joints, translation, shape, translation velocity, and translation acceleration with weights $1$, $50$, $1$, $0.5$, $2$, and $0.5$, respectively. We train for 100K iterations using AdamW with batch size 32,learning rate $10^{-4}$ for the mapping prior, and $10^{-5}$ for the signal encoder.

\paragraph{Real-world adaptation and inference.}
The pretrained signal-encoder weights remain frozen except for LoRA modules inserted into its attention and feed-forward layers (rank $r=16$, LoRA alpha $\alpha=32$, dropout $0.1$). The mapping modules and prediction heads remain trainable, while the motion tokenizer is frozen. AdamW uses learning rates $10^{-4}$ for the mapping modules and $3\times10^{-5}$ for LoRA, weight decay $10^{-4}$, and effective batch size eight. For Cls with motion-prior-guided adaptation, the frozen tokenizer converts reference rotations into code targets, and the token cross-entropy weight is linearly warmed to $0.05$ over the first 500 steps.

mmSimPrior-Real adaptation uses 24 support sequences, a fixed 4K-iteration horizon, and seed 2026.
RT-Pose uses the same support budget and effective batch size, with a common 8K horizon for all methods.
All reported checkpoints are evaluated over all 32 target frames without target-test checkpoint selection.
On mmSimPrior-Real, reported Reg results include evaluation-time motion-prior refinement (MPR) unless direct and refined outputs are explicitly compared.

\paragraph{Physics-informed domain randomization.}
Propagation-level perturbations comprise multipath ghosting and spatial response spreading, while acquisition-level perturbations comprise response-statistics alignment and spatial-frequency filtering. Ghosting is applied to the bird's-eye and side views with probability $0.5$.
Gaussian spreading is applied to the front view with probability $0.5$. The perturbation strength is increased progressively during pretraining.

Response alignment performs per-frame histogram matching to fixed per-modality reference histograms estimated once from an unlabeled Hall recording that does not appear in any evaluation manifest. The histograms contain radar intensities only; no RT-Pose data or pose,
mesh, or motion-category annotations are used. Alignment is applied independently to each modality with probability $0.5$. Spatial-frequency filtering uses a radial low-pass mask with cutoff $0.15$ cycles/pixel and
probabilities $0.3$ during signal-prior pretraining and $0.5$ during mapping pretraining. Random strengths are shared across frames of a sequence to preserve temporal consistency. All perturbations are disabled during real-world adaptation and evaluation, and the same configuration is fixed across benchmarks.

All models were implemented in PyTorch and trained on NVIDIA A6000 GPUs.
\section{Radar Simulation and Signal Processing}
\label{supp:radar_simulation}

\subsection{IF Signal Simulation from Dynamic Human Meshes}
\label{supp:if_simulation}

We generate mmSimPrior-Sim by driving a physics-based FMCW radar simulator with AMASS motion sequences. Each sequence is represented as a temporally varying SMPL-X mesh and resampled to 10 FPS. At every radar frame, mesh facets are characterized by their barycenters, surface normals, and areas. Back-facing and occluded facets are removed using viewpoint-conditioned visibility filtering, including Hidden Point Removal over the facet locations.

For each retained facet and transmit--receive antenna pair, the simulator computes a bistatic propagation path over the sampled frequencies.
Reflection strength depends on the transmit--facet and facet--receive path lengths, phase delay, geometric attenuation, facet area, and surface orientation. The facet responses are coherently accumulated to synthesize the complex IF signal according to the bistatic reflection model in the main paper.

Each AMASS motion is rendered at three radar--subject distances, introducing variation in attenuation, angular resolution, and the spatial distribution of human reflections. The resulting complex IF sequences are paired with aligned SMPL-X pose, root translation, and body-shape annotations. The simulated radar observations supervise the signal and mapping priors, while the corresponding AMASS trajectories are also used to train the motion prior.

\subsection{Heatmap Construction from IF Signals}
\label{supp:prelim_radar}

The simulated and real IF signals are converted into a common four-modality radar representation consisting of one Range--Doppler heatmap and three orthogonal MVDR spatial heatmaps.
Applying the same signal-processing pipeline to simulation, mmSimPrior-Real, and RT-Pose prevents representation-specific differences from confounding the transfer evaluation.

\paragraph{Range--Doppler heatmap.}
Let
$\mathbf{X}_t \in
\mathbb{C}^{N_{\mathrm{tx}}\times N_{\mathrm{rx}}
\times N_{\mathrm{c}}\times N_{\mathrm{s}}}$
denote the IF tensor at radar frame $t$, where
$N_{\mathrm{tx}}$ and $N_{\mathrm{rx}}$ are the numbers of transmit
and receive antennas,
$N_{\mathrm{c}}$ is the number of chirps, and
$N_{\mathrm{s}}$ is the number of ADC samples per chirp.
After arranging the transmit--receive pairs into
$N_{\mathrm{ant}}=N_{\mathrm{tx}}N_{\mathrm{rx}}$
virtual antenna channels, we apply a windowed Range-FFT along the
fast-time dimension and a Doppler-FFT along the slow-time dimension.
The resulting Range--Doppler response is

\begin{equation}
\mathbf{H}^{\mathrm{RD}}_t[k,f]
=
\left|
\frac{1}{N_{\mathrm{ant}}}
\sum_{a=1}^{N_{\mathrm{ant}}}
\sum_{l=0}^{N_{\mathrm{c}}-1}
\mathbf{Z}_t[a,l,k]
e^{-j2\pi fl/N_{\mathrm{c}}}
\right|,
\label{supp:eq_rd_fft}
\end{equation}

where $\mathbf{Z}_t$ is the range-domain response and $k$ and $f$
index the range and Doppler bins.
The RD modality preserves radial-velocity evidence complementary to
the spatial MVDR views.

\paragraph{Multi-view MVDR spatial heatmaps.}
For each range bin, we estimate the antenna-domain covariance and
evaluate the MVDR spectrum $P_t(\mathbf{p})$ over a discretized
3D spatial grid $\mathbf{p}=(x,y,z)$.
Rather than directly processing the full response volume, we compute
three orthogonal mean projections:

\begin{equation}
\begin{aligned}
\mathbf{H}^{\mathrm{front}}_t(x,z)
    &= \mathbb{E}_{y}\!\left[P_t(x,y,z)\right],\\
\mathbf{H}^{\mathrm{bird}}_t(x,y)
    &= \mathbb{E}_{z}\!\left[P_t(x,y,z)\right],\\
\mathbf{H}^{\mathrm{side}}_t(y,z)
    &= \mathbb{E}_{x}\!\left[P_t(x,y,z)\right].
\end{aligned}
\label{supp:eq_mvdr_views}
\end{equation}

The front and side views have spatial resolution $128\times50$, while the bird's-eye and RD heatmaps have resolution $128\times128$. All modalities are log-compressed and independently normalized per frame.
Together, the RD heatmap and three MVDR projections form the common four-modality observation used for simulated pretraining, real-world adaptation, and evaluation.
\begin{table*}[t]
\centering
\small
\setlength{\tabcolsep}{3pt}
\renewcommand{\arraystretch}{1.15}
\renewcommand{\arraystretch}{1.2}
\begin{tabular*}{\textwidth}{@{\extracolsep{\fill}} c l l c c c c c @{}}
\toprule
\textbf{Task} & \textbf{Dataset} & \textbf{Data} & \textbf{\#Subj} & \textbf{\#Seq} & \textbf{\#Frame} & \textbf{Sim} & \textbf{Public}\\
\midrule

\multirow{3}{*}{\centering\makecell{HMR}}
& MPI-INF-3DHP~\cite{mehta2017monocular} & RGB & 8 & 16 & 1.3M & $\times$ & \checkmark \\
& Human3.6M~\cite{ionescu2013human3} & RGB & 11 & 210 & 3.6M & $\times$ & \checkmark\\

& BEDLAM2.0~\cite{tesch2025bedlam2} & RGB & $>$4K & $>$27K & $>$8M & \checkmark & \checkmark \\

\midrule
\multirow{7}{*}{\makecell[c]{HPE}}
& RadHAR~\cite{singh2019radhar} & PC & 2 & 16 & 167K & $\times$ & \checkmark\\
& mm-Pose~\cite{mmPose} & PC & 2 & 8 & 40K & $\times$ & $\times$ \\
& MARS~\cite{an2021mars} & PC & 4 & 80 & 40K & $\times$ & \checkmark \\
& mRI~\cite{an2022mri} & PC & 20 & 300 & 160K & $\times$ & $\times$ \\
& MM-Fi~\cite{yang2023mm} & PC & 40 & 1K & 320K & $\times$ & \checkmark\\
& RF-Pose~\cite{zhao2018through} & HM & 100 & -- & -- & $\times$ & $\times$ \\
& HuPR~\cite{lee2023hupr} & HM & 6 & 235 & 141K & $\times$ & \checkmark\\

\midrule
\multirow{9}{*}{\makecell[c]{HMR}}
& mmMesh~\cite{xue2021mmmesh} & PC & 20 & 160 & 480K
& $\times$ & $\times$ \\
& mmBody~\cite{chen2022mmbody} & PC & 20 & 48 & $<$70K
& $\times$ & \checkmark \\
& HIBER~\cite{wu2022rfmask} & HM & -- & -- & 179K
& $\times$ & Partial \\
& MMVR~\cite{rahman2024mmvr} & HM & 20 & 395 & 345K
& $\times$ & \checkmark \\
& RT-Pose~\cite{ho2024rt} & HM & -- & 240 & 72K
& $\times$ & \checkmark \\
& RF-Genesis~\cite{chen2023rf} & PC & -- & -- & --
& \checkmark & $\times$ \\
& mmGPE~\cite{xue2023towards} & HM & 4 & 34 & 276K
& \checkmark & $\times$ \\
\cmidrule(lr){2-8}
& \textbf{mmSimPrior-Sim (Ours)}
& \textbf{HM} & \textbf{344} & \textbf{30K} & \textbf{4.0M}
& \checkmark & \checkmark$^{*}$ \\
& \textbf{mmSimPrior-Real (Ours)}
& \textbf{HM} & \textbf{6} & \textbf{1,126} & \textbf{162K}
& $\times$ & \checkmark$^{*}$ \\

\bottomrule
\end{tabular*}
\caption{
\textbf{Comparison of RGB- and radar-based human motion datasets
by task, sensing representation, and scale.}
HMR denotes human motion reconstruction, PC denotes mmWave radar
point clouds, and HM denotes radar heatmaps.
Sequence counts follow the definitions of the original datasets and
are therefore not directly comparable across datasets.
\textbf{mmSimPrior-Sim} provides large-scale paired radar--motion
supervision generated from AMASS, whereas
\textbf{mmSimPrior-Real} provides a real-world benchmark for
zero-shot and limited-data adaptation.
Together, they form the mmSimPrior dataset suite with 4.2M frames
and 31K sequences.
$^{*}$To be released upon acceptance.
}
\label{supp:tab_dataset_comparison}
\end{table*}

\begin{figure*}[ht]
  \centering
  \includegraphics[width=0.97\linewidth]{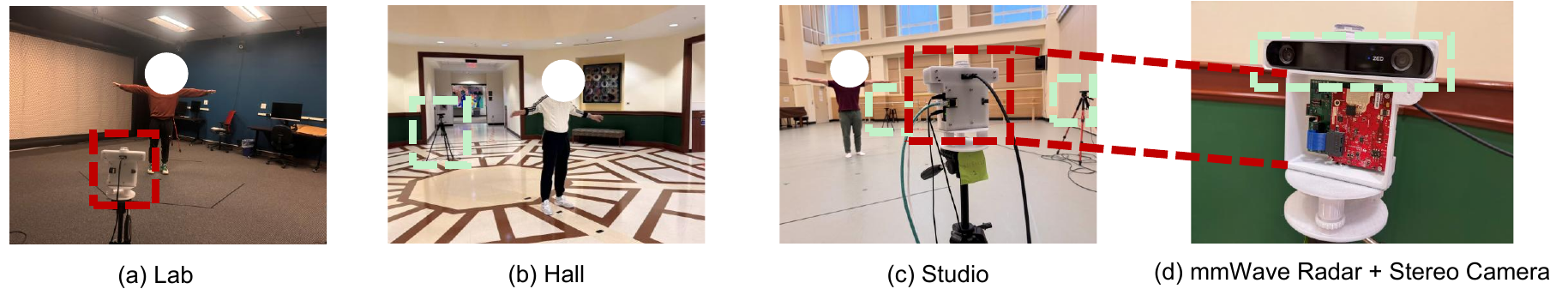}
  \caption{
\textbf{Real-world data collection setup.}
We collect synchronized mmWave radar and stereo-camera observations
in three indoor environments:
(a) Lab, (b) Hall, and (c) Studio.
The environments differ in spatial layout, clutter, and reflective
structures, producing distinct multipath and sensing conditions.
(d) The radar and stereo camera are rigidly mounted on the same tripod.
Stereo images are used only to construct reference SMPL-X annotations;
all evaluated models use radar heatmaps alone.
}
  \label{supp:fig_data_collection}
\end{figure*}

\begin{figure*}[ht]
    \centering
  \includegraphics[width=0.97\linewidth]{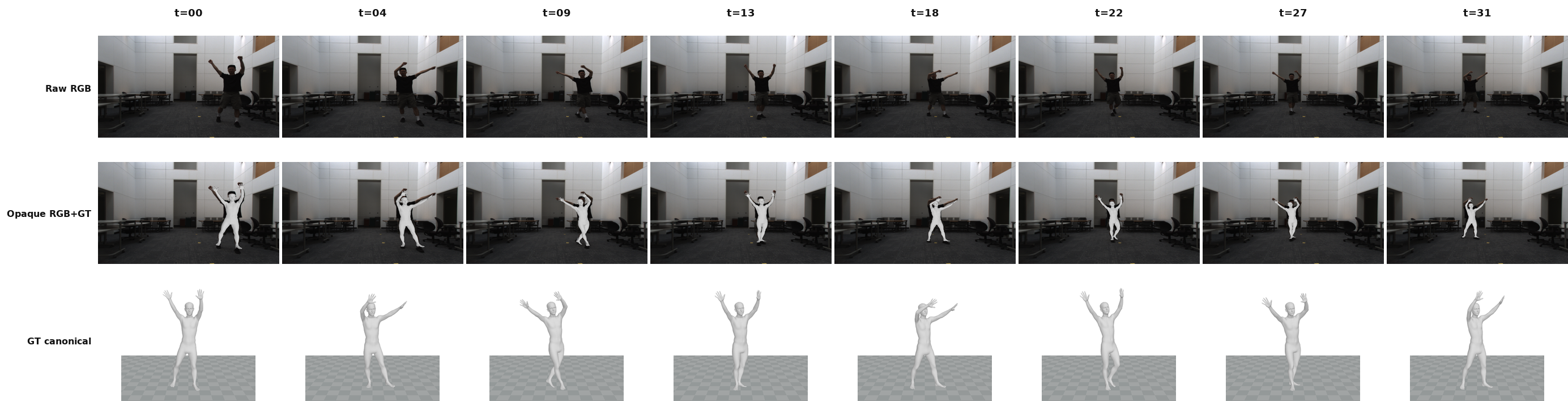}
  \caption{
\textbf{Pseudo-SMPL-X annotation examples on RT-Pose.} Rows show the raw RGB observation, the fitted SMPL-X overlay, and the canonicalized reference mesh across the sequence.
}
\label{supp:pseudo_rtpose}
\end{figure*}

\section{Dataset and Evaluation Protocol}
\label{supp:dataset_protocol}

\paragraph{mmSimPrior dataset suite.}
The mmSimPrior dataset suite consists of
\textbf{mmSimPrior-Sim}, a large-scale simulated pretraining corpus,
and \textbf{mmSimPrior-Real}, a paired real-world benchmark for
evaluating zero-shot and limited-data transfer.
mmSimPrior-Sim is generated from AMASS motions spanning
344 mocap subjects and contains approximately 4.0M frames
from 30K radar--motion sequences.
mmSimPrior-Real contains 1,126 synchronized radar--motion recordings
collected from six participants across three indoor environments,
40 motion categories, and three sensing-distance ranges.
Together, the two components contain 4.2M frames, 31K sequences, and 350 subjects.

\paragraph{mmSimPrior-Real collection.}
As illustrated in Fig.~\ref{supp:fig_data_collection}, we collect synchronized mmWave radar and stereo-camera observations in three indoor environments: Hall, Lab, and Studio. 
The environments differ in room geometry, background clutter, reflective surfaces, and radar placement, resulting in distinct multipath and acquisition conditions.
The radar and stereo camera are rigidly mounted on the same tripod and temporally synchronized.
We use a Texas Instruments AWR1843BOOST FMCW radar configured with three transmit and four receive channels. Raw complex ADC data are captured through a DCA1000EVM at 10 FPS, using 128 chirps per frame
and 256 ADC samples per chirp, and are processed into one RD and three orthogonal MVDR heatmaps following Sec.~\ref{supp:prelim_radar}. 
The stereo observations are used only to construct reference SMPL-X annotations; all reconstruction models receive radar heatmaps alone during adaptation and evaluation.

The complete benchmark contains 1,126 unique recordings.
For each environment, 24 recordings form a fixed support set for
limited-data adaptation, while the remaining 1,054 recordings form
the evaluation population.
The support and evaluation recordings are sequence-disjoint.

All data-collection procedures involving human participants were
reviewed and approved by our Institutional Review Board.
Each participant provided informed consent and was informed of the
sensing devices, collection procedure, and intended research use.

\subsection{Evaluation Protocol Details}
\label{supp:evaluation_protocol}

\paragraph{Metrics.}
We report MPJPE over the 22 SMPL-X body joints and PA-MPJPE after framewise similarity alignment.
W-MPJPE evaluates absolute body placement in the world coordinate system, while MPVPE measures the mean error over the SMPL-X body vertices.
Because the real-world mesh references are obtained through camera-based reconstruction and fitting, MPVPE is interpreted as a mesh-level reconstruction metric rather than an isolated measure of body-shape accuracy.
All errors are reported in millimeters.

\paragraph{Averaging.}
Unless otherwise specified, results are averaged uniformly over valid test recordings.
Each recording therefore contributes equally, irrespective of its number of frames.
When results are grouped by environment, sensing distance, or motion category, the reported aggregate is weighted by the number of valid recordings in each group.

\paragraph{NOS split construction.}
For each real-world recording, we define its sensing configuration by four factors: subject identity, environment, sensing location, and
motion category.
The No-Overlap Setting (NOS) requires that no adaptation--test pair
share the same complete four-factor configuration.
Individual factors may overlap, but every test recording must differ from every adaptation recording in at least one factor.
This prevents evaluation on another temporal segment of an identical subject--environment--location--motion configuration.

For same-environment adaptation, we use 24 recordings from the target environment and evaluate on the remaining recordings satisfying NOS.
For cross-environment adaptation, the adaptation and evaluation environments differ, and the source--target direction is specified in the corresponding table.
Zero-shot denotes evaluation without pose annotations, paired radar--motion supervision, adaptation, or model selection from the evaluated population. The response-alignment histograms are estimated once from a separate unlabeled Hall recording excluded from the evaluation manifest and are fixed for all experiments.

\paragraph{Adaptation-set duration.}
The fixed 24-recording support sets contain 7,472, 2,351, and
3,479 supervised frames for Hall, Lab, and Studio, respectively,
corresponding to approximately 12.5, 3.9, and 5.8 minutes at 10\,Hz.
For RT-Pose, the fixed support set contains 24 32-frame sequences,
or 768 frames (76.8 seconds) in total.

\paragraph{Baseline adaptation.}
The compared methods were originally developed with different radar
representations and temporal protocols.
For a controlled comparison, we adapt each baseline to the common
four-modality observation consisting of one Range--Doppler heatmap
and three orthogonal MVDR views, while preserving its defining
temporal modeling and pose-prediction mechanism.
All methods use the same simulation corpus, 24-sequence real-data
budget, fixed adaptation horizon, NOS split, and evaluation metrics.

Evaluation covers the same 32-frame target window at 10 FPS.
Methods with a shorter native temporal context process the window in
contiguous segments, and metrics are aggregated over all 32 target
frames.
The $^\dagger$ marker denotes baseline variants pretrained under this
unified simulated-pretraining and evaluation protocol.

\subsection{External Validation on RT-Pose}
\label{supp:rtpose_processing}

We additionally evaluate on the public RT-Pose benchmark
\cite{ho2024rt}, which provides synchronized radar ADC measurements,
radar tensors, stereo RGB observations, LiDAR point clouds, and
3D body-joint annotations across diverse indoor and outdoor scenes.
We focus on its single-person recordings.
Starting from the released ADC signals, we apply the same
signal-processing pipeline used for mmSimPrior-Real to generate one
Range--Doppler heatmap and three orthogonal MVDR views.
All compared methods therefore receive the same four-modality radar
input and are evaluated over 32-frame windows at 10 FPS. 
For limited-data adaptation, we use a fixed 24-sequence support set and report results on 111 separately screened evaluation sequences. The support and evaluation sets are completely sequence-disjoint, with no shared recordings or frames.

\paragraph{Pseudo-SMPL-X reference validation.}
We assess the fitted references on 12 common limb landmarks: the bilateral
hips, knees, ankles, shoulders, elbows, and wrists. Because the released 3D
keypoints and left-view 2D detections participate in fitting, these diagnostics
characterize fitting and image-level consistency. Lower-body joints differ
from the released 3D keypoints by 39.6\,mm on average, with hip PCK@50 above
80\%. Median reprojection errors are 17.0 pixels in the fitting view and
24.1 pixels in the held-out right view. The larger upper-body residual
(140.8\,mm) primarily reflects the different shoulder definitions used by
RT-Pose surface landmarks and SMPL-X kinematic joint centers. We use the fitted
references for shared comparative mesh evaluation and additionally report
evaluation against the official RT-Pose keypoints in Sec.~D.8, which preserves
the method ranking and a similar Reg margin.

Temporal regularization produces consistent pose, shape, and root-translation
trajectories. Before evaluating any radar method, we exclude sequences with
severe occlusion, low illumination, or fitting divergence using only the RGB
annotation pipeline; 111 sequences pass this screening. RGB is used only for
offline reference construction, while adaptation and evaluation use radar
heatmaps alone.
\section{Additional Experimental Results}
\label{supp:additional_results}

We complement the main evaluation with analyses of data efficiency, held-out
factors, mesh and temporal quality, and tokenizer capacity. Unless otherwise
specified, errors are reported in millimeters and lower values are better. The
best and second-best results are shown in \textbf{boldface} and
\underline{underlined}, respectively. All experiments follow
Sec.~\ref{supp:evaluation_protocol}. On mmSimPrior-Real, mmSimPrior-Reg includes
evaluation-time motion-prior refinement (MPR) unless direct and refined outputs
are explicitly compared.

% ============================================================
\subsection{Data-Efficient Real-World Adaptation}
\label{supp:fewshot_curve}

We evaluate mmSimPrior-Reg and mmSimPrior-Cls with 0, 1, 4, 8, and 24 real-world
adaptation sequences; 0 denotes zero-shot evaluation. The same-environment
setting adapts and evaluates in Hall under NOS, whereas the cross-environment
setting adapts on Hall and evaluates on Lab. The three seeds use different support
subsets at these lower budgets, while all runs use the same complete
24-sequence support set at the largest budget. Every nonzero-budget run
uses the fixed 4,000-step adaptation schedule.

\begin{figure}[tbp]
    \centering
    \includegraphics[width=\columnwidth]{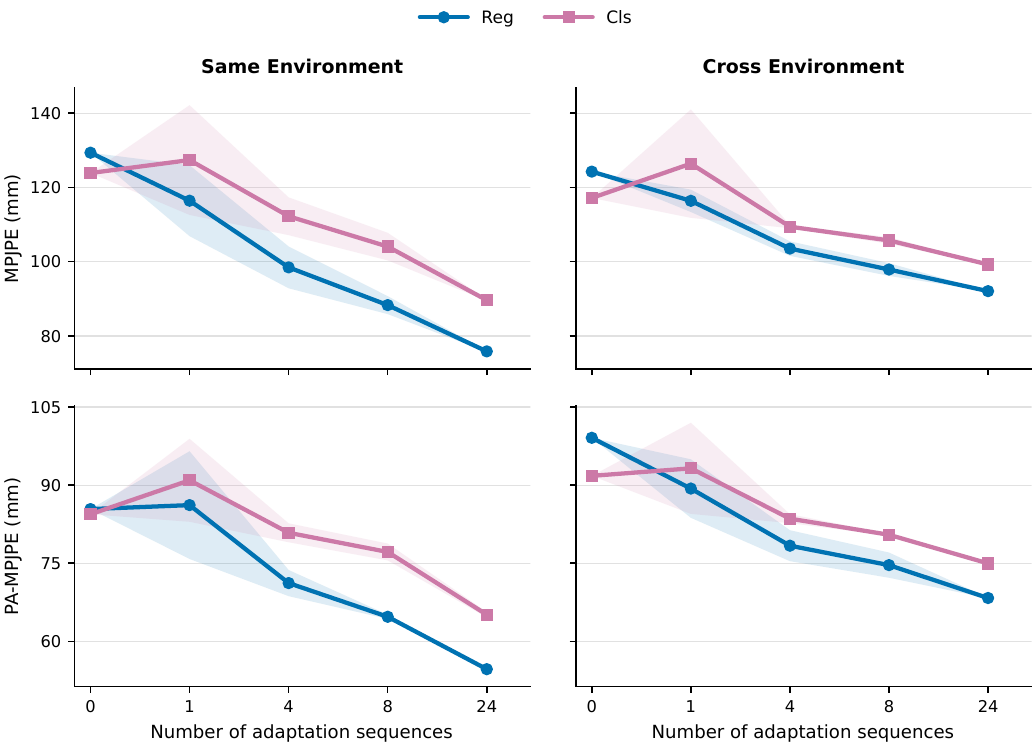}
    \caption{\textbf{Few-shot adaptation on mmSimPrior-Real.}
    The left column reports same-environment Hall adaptation and the right
    column reports Hall$\rightarrow$Lab cross-environment transfer, evaluated
    on fixed test sets containing 89 and 424 sequences, respectively. The top
    and bottom rows show MPJPE and PA-MPJPE. Curves report the mean over three
    deterministic runs (seeds 2026--2028), and shaded regions show $\pm1$
    standard deviation.}
    \label{supp:fig_fewshot_curve}
\end{figure}

Fig.~\ref{supp:fig_fewshot_curve} shows complementary behavior. Cls provides the
stronger Hall-to-Lab zero-shot initialization, whereas Reg improves faster once
several labeled sequences are available and achieves the lowest final errors.
The one-sequence Cls result shows a small non-monotonic variation; the trend
becomes consistent from four sequences onward.

\begin{table}[t]
\centering
\footnotesize
\setlength{\tabcolsep}{4.0pt}
\renewcommand{\arraystretch}{1.12}
\caption{\textbf{Stability of limited-data adaptation.}
mmSimPrior-Reg under three adaptation runs with the simulation
checkpoint, 4K horizon, and test manifest held fixed. Lab and Studio
vary only the optimization seed; Hall additionally draws a different
24-sequence support subset per run. Values are mean $\pm$ sample
standard deviation in millimeters.}
\label{supp:tab_adaptation_stability}
\begin{tabular}{@{}l|ccc@{}}
\toprule
Metric & Hall$^{\ast}$ & Lab & Studio \\
\midrule
MPJPE$\downarrow$    & $75.8\pm0.1$  & $77.2\pm0.0$  & $62.1\pm0.1$ \\
PA-MPJPE$\downarrow$ & $54.7\pm0.0$  & $57.0\pm0.0$  & $49.5\pm0.0$ \\
W-MPJPE$\downarrow$  & $115.7\pm0.4$ & $116.1\pm0.1$ & $94.6\pm0.3$ \\
MPVPE$\downarrow$    & $101.8\pm0.2$ & $103.5\pm0.0$ & $85.3\pm0.0$ \\
\bottomrule
\end{tabular}

\vspace{2pt}
{\scriptsize $^{\ast}$Support subset also varies across runs.}
\end{table}

\paragraph{Adaptation stability.}
Because limited-data adaptation uses only 24 support sequences, we
verify that the reported results are not an artifact of a particular
adaptation run. Table~\ref{supp:tab_adaptation_stability} repeats the
24-sequence adaptation three times per environment. Every resulting
checkpoint differs, but the errors do not: the largest standard
deviation across all environments and metrics is $0.36$~mm, two orders
of magnitude below the $24.7$--$39.0\%$ margins to the strongest adapted baseline. 
In Hall the support subset varies across runs as well, so the same conclusion holds for support selection
and not only for optimization noise. mmSimPrior-Cls is comparably
stable, with a maximum standard deviation of $0.4$~mm across Lab and
Studio. The main tables report the seed-2026 run, and all baselines use
the same fixed seed.

\begin{table*}[tbp]
\centering
\footnotesize
\setlength{\tabcolsep}{3.0pt}
\renewcommand{\arraystretch}{1.05}

\textbf{(a) Motion categories unseen during real-world adaptation}

\vspace{3pt}

\begin{tabular}{l|cccc}
\toprule
Method & Hall & Lab & Studio & Avg. \\
\midrule
mmDiff$^\dagger$~\cite{fan2024diffusion} & 109.6 / 73.4 & 107.5 / 73.9 & 109.4 / 73.7 & 108.7 / 73.7 \\
RAPTR$^\dagger$~\cite{kato2025raptr}  & 121.2 / 85.3 & 112.2 / 81.5 & 123.4 / 89.8 & 118.8 / 86.0 \\
mmGPE$^\dagger$~\cite{xue2023towards} & 112.1 / 73.4 & 110.2 / 72.9 & 106.9 / 70.2 & 108.9 / 71.7 \\
\midrule
\textbf{mmSimPrior-Reg}
& \textbf{85.8} / \textbf{58.4}
& \textbf{87.4} / \textbf{59.8}
& \textbf{68.4} / \textbf{51.8}
& \textbf{78.0} / \textbf{55.8} \\
\textbf{mmSimPrior-Cls}
& \underline{100.0} / \underline{68.6}
& \underline{100.2} / \underline{69.0}
& \underline{93.0} / \underline{64.5}
& \underline{96.7} / \underline{66.7} \\
\bottomrule
\end{tabular}

\vspace{6pt}

\textbf{(b) Per-environment zero-shot transfer}

\vspace{3pt}

\begin{tabular}{l|cccc}
\toprule
Method & Hall & Lab & Studio & Avg. \\
\midrule
mmDiff$^\dagger$~\cite{fan2024diffusion} & 139.2 / 107.0 & 133.9 / 104.1 & 125.4 / 90.2 & 130.0 / 97.2 \\
RAPTR$^\dagger$~\cite{kato2025raptr}  & 187.7 / 103.2 & 175.5 / 114.7 & 163.2 / 95.8 & 170.2 / 104.0 \\
mmGPE$^\dagger$~\cite{xue2023towards} & 156.0 / 100.8 & 142.8 / 99.7 & 137.7 / 94.2 & 141.3 / 97.0 \\
\midrule
\textbf{mmSimPrior-Reg}
& \underline{129.4} / \underline{85.4}
& \underline{124.3} / \underline{99.1}
& \textbf{113.3} / \underline{89.7}
& \underline{119.1} / \underline{93.1} \\
\textbf{mmSimPrior-Cls}
& \textbf{123.9} / \textbf{84.4}
& \textbf{117.2} / \textbf{91.8}
& \underline{119.4} / \textbf{88.9}
& \textbf{118.9} / \textbf{89.7} \\
\bottomrule
\end{tabular}

\caption{\textbf{Generalization across held-out motions and environments.}
Panel (a) evaluates motion categories absent from the 24-sequence adaptation set
on 59 Hall, 175 Lab, and 220 Studio recordings. Panel (b) reports zero-shot
performance over 89 Hall, 424 Lab, and 541 Studio recordings. Each entry is
MPJPE / PA-MPJPE; Avg. is sequence-weighted.}
\label{supp:tab_generalization_breakdown}
\label{supp:tab_unseen_pose}
\label{supp:tab_zeroshot_per_env}
\end{table*}

% ============================================================
\subsection{Generalization to Held-Out Factors}
\label{supp:heldout_generalization}

We isolate three sources of real-world shift: subject identity, motion category,
and environment. The tested factor is absent from adaptation, or the entire
target benchmark is unseen in the zero-shot setting. Avg. is computed from unrounded sequence-level errors and is
weighted by the number of valid recordings.

\paragraph{Held-out subjects.}
Table~\ref{supp:tab_unseen_subject} reports nine-fold leave-one-subject-out
adaptation. Four Lab folds and five Studio folds reserve one participant for
testing while adapting on 24 sequences from the remaining participants. Hall is
excluded because it contains only one participant.

\begin{table}[tbp]
\centering
\footnotesize
\setlength{\tabcolsep}{1.6pt}
\renewcommand{\arraystretch}{1.12}
\caption{\textbf{Nine-fold leave-one-subject-out adaptation.}
Results are sequence-weighted over 1,013 held-out-subject recordings.}
\label{supp:tab_unseen_subject}
\begin{tabular}{l|cccc}
\toprule
Method & MPJPE$\downarrow$ & PA-MPJPE$\downarrow$ & W-MPJPE$\downarrow$ & MPVPE$\downarrow$ \\
\midrule
mmDiff$^\dagger$ & 105.6 & 81.6 & 172.4 & 150.9 \\
RAPTR$^\dagger$  & 117.3 & 92.7 & 176.8 & 168.6 \\
mmGPE$^\dagger$  & 106.5 & 77.2 & 142.3 & 148.4\\
\midrule
\textbf{mmSimPrior-Reg} & \textbf{72.6} & \textbf{54.6} & \textbf{108.0} & \textbf{98.0} \\
\textbf{mmSimPrior-Cls} & \underline{92.4} & \underline{69.4} & \underline{124.4} & \underline{126.4} \\
\bottomrule
\end{tabular}
\end{table}

Across the nine folds, Reg and Cls obtain unweighted fold-level MPJPEs of
$72.7\pm8.6$ and $92.4\pm4.9$~mm. The sequence-weighted results confirm that
both modes transfer to unseen body geometry and motion style, with Reg providing
the strongest accuracy.

\paragraph{Held-out motions and environments.}
Table~\ref{supp:tab_generalization_breakdown} removes motion categories from the
adaptation set and separately evaluates each environment without any target-
benchmark adaptation or model selection.

Reg is strongest in every environment when limited adaptation is allowed. In
zero-shot transfer, Cls is strongest in Hall and Lab and gives the best weighted
average, supporting the discrete prior when target supervision is unavailable.

% ============================================================
\subsection{Generalization to Distance and Motion Regime}
\label{supp:factorwise_robustness}

We further examine whether the adapted models remain robust to
variation in sensing distance and motion regime.
Because zero-shot behavior is already analyzed in
Fig.~\ref{supp:fig_fewshot_curve} and
Table~\ref{supp:tab_generalization_breakdown},
we focus here on MPJPE after 24-sequence real-world adaptation.
\begin{table}[t]
\centering
\footnotesize
\setlength{\tabcolsep}{3.0pt}
\renewcommand{\arraystretch}{1.05}
\caption{
\textbf{MPJPE across sensing distances after 24-sequence adaptation.}
Near, Mid, and Far contain 303, 379, and 372 recordings,
respectively. Avg. is sequence-weighted.
}
\label{supp:tab_distance_gen}
\begin{tabular}{@{}l|cccc@{}}
\toprule
Method & Near & Mid & Far & Avg. \\
\midrule
mmDiff$^\dagger$~\cite{fan2024diffusion}
& 95.1 & 104.9 & 109.8 & 103.8 \\
RAPTR$^\dagger$~\cite{kato2025raptr}
& 107.5 & 115.3 & 117.6 & 113.9 \\
mmGPE$^\dagger$~\cite{xue2023towards}
& 97.5 & 102.0 & 109.7 & 103.4 \\
\midrule
\textbf{mmSimPrior-Reg} 
& \textbf{65.7}
& \textbf{68.3}
& \textbf{73.4} 
& \textbf{69.4} \\
\textbf{mmSimPrior-Cls}
& \underline{84.3}
& \underline{87.1}
& \underline{94.7}
& \underline{89.0} \\
\bottomrule
\end{tabular}
\end{table}

\begin{table}[t]
\centering
\footnotesize
\setlength{\tabcolsep}{4.0pt}
\renewcommand{\arraystretch}{1.05}
\caption{
\textbf{MPJPE across motion regimes after 24-sequence adaptation.}
The evaluation set contains 872 in-place and 182 locomotion
recordings. Avg. is sequence-weighted.
}
\label{supp:tab_motion_regime}
\begin{tabular}{@{}l|ccc@{}}
\toprule
Method & In-place & Locomotion & Avg. \\
\midrule
mmDiff$^\dagger$~\cite{fan2024diffusion}
& 95.0 & 146.1 & 103.8 \\
RAPTR$^\dagger$~\cite{kato2025raptr}
& 105.6 & 153.2 & 113.9 \\
mmGPE$^\dagger$~\cite{xue2023towards}
 & 96.8 & 135.3 & 103.4 \\
\midrule
\textbf{mmSimPrior-Reg}
& \textbf{66.5} & \textbf{83.1} & \textbf{69.4}\\
\textbf{mmSimPrior-Cls}
& \underline{83.9}
& \underline{113.4}
& \underline{89.0} \\
\bottomrule
\end{tabular}
\end{table}
Tables~\ref{supp:tab_distance_gen} and
\ref{supp:tab_motion_regime} provide complementary diagnostics
of the adapted models.
Across sensing distances, all methods exhibit some degradation as
the subject moves farther from the radar.
Nevertheless, mmSimPrior-Reg remains strongest in every range,
with MPJPE increasing moderately from 65.7~mm in the Near group to 73.4~mm in the Far group.
mmSimPrior-Cls consistently ranks second, indicating that both
prediction modes retain their advantage under reduced signal
strength and spatial resolution.

Motion regime produces a larger performance difference.
For mmSimPrior-Reg, MPJPE increases from 66.5~mm for in-place motions to 83.1~mm for locomotion, while mmSimPrior-Cls increases from 83.9~mm to 113.4~mm.
Despite this more challenging setting, both variants maintain clear
margins over all adapted baselines.
These results suggest that the learned simulation priors remain
robust to sensing-range variation, while locomotion remains the
more difficult real-world operating regime.

% ============================================================
\subsection{Mesh Accuracy and Temporal Refinement}
\label{supp:mesh_temporal_quality}

We evaluate mesh-level reconstruction and isolate the effect of projecting direct
Reg predictions through the frozen motion prior.

\begin{table}[tbp]
\centering
\small
\renewcommand{\arraystretch}{1.12}
\caption{\textbf{Mesh-level reconstruction.}
Sequence-weighted MPVPE over the common 1,054-sequence population.}
\label{supp:tab_mpvpe_mesh}
\label{supp:tab_mesh_temporal}
\begin{tabular}{l|cc}
\toprule
Method & Zero-shot & 24-sequence \\
\midrule
mmDiff$^\dagger$~\cite{fan2024diffusion} & 189.7 & 148.5 \\
RAPTR$^\dagger$~\cite{kato2025raptr}  & 244.1 & 163.6 \\
mmGPE$^\dagger$~\cite{xue2023towards} & 207.2 & 145.9 \\
\midrule
\textbf{mmSimPrior-Reg} & \underline{173.4} & \textbf{94.0} \\
\textbf{mmSimPrior-Cls} & \textbf{172.4} & \underline{121.2} \\
\bottomrule
\end{tabular}
\end{table}

\begin{table}[tbp]
\centering
\footnotesize
\setlength{\tabcolsep}{7pt}
\renewcommand{\arraystretch}{1.1}
\caption{\textbf{Effect of motion-prior refinement.}
Relative change is measured from Direct Reg; negative is better.}
\label{supp:tab_mpr_ablation}
\begin{tabular}{l|rrr}
\toprule
Metric & Direct & +MPR & Rel. change \\
\midrule
MPJPE$\downarrow$             & 68.6 & 69.4 & +1.1\% \\
PA-MPJPE$\downarrow$          & 52.7 & 53.0 & +0.5\%  \\
W-MPJPE$\downarrow$           & 104.7 & 105.2 & +0.4\% \\
MPVPE$\downarrow$             & 92.8 & 94.0 & +1.3\% \\
\midrule
Joint Vel. Err.$\downarrow$ & 41.7 & 38.5 & \textbf{-7.7\%} \\
Joint Accel. Err.$\downarrow$ & 49.9 & 40.9 & \textbf{-18.0\%} \\
Pred. Jerk$\downarrow$ & 63.1 & 40.0  & \textbf{-36.7\%} \\
\bottomrule
\end{tabular}
\end{table}

Cls gives the best zero-shot MPVPE, whereas Reg is substantially stronger after
adaptation. MPR changes frame-level spatial errors by at most 1.3\%, but reduces
velocity, acceleration, and jerk errors by 7.7\%, 18.0\%, and 36.7\%,
respectively. It therefore acts primarily as a temporal regularizer.

% ============================================================
\subsection{Motion-Tokenizer Capacity}
\label{supp:vq_ablation}

Finally, we test whether the frozen tokenizer imposes a reconstruction ceiling by
varying its codebook and number of joint-temporal tokens.

\begin{table}[htbp]
\centering
\scriptsize
\setlength{\tabcolsep}{7pt}
\renewcommand{\arraystretch}{1.12}
\caption{\textbf{Motion-tokenizer capacity.}
Each entry is MPJPE / PA-MPJPE / MPVPE. The codebook block fixes 320 tokens;
the token-count block fixes a $2048\times256$ codebook.}
\label{supp:tab_vq_ablation}
\begin{tabular}{@{}ll|cc@{}}
\toprule
& Setting & AMASS & Lab \\
\midrule
\multirow{3}{*}{Codebook}
& $1024\times256$ & 10.45 / 6.28 / 13.90 & 14.24 / 9.55 / 19.43 \\
& $2048\times128$ & 10.11 / 5.42 / 13.30 & 13.60 / 8.75 / 18.18 \\
& $2048\times256$ & \textbf{9.15 / 5.39 / 12.26} & \textbf{10.38 / 7.95 / 14.54} \\
\midrule
\multirow{3}{*}{Tokens}
& 80  & 20.47 / 12.79 / 28.13 & 30.83 / 21.44 / 43.65 \\
& 160 & 14.95 / 9.06 / 20.19 & 21.99 / 14.84 / 30.07 \\
& 320 & \textbf{9.15 / 5.39 / 12.26} & \textbf{10.38 / 7.95 / 14.54} \\
\bottomrule
\end{tabular}
\end{table}

At the selected $2048\times256$ codebook with 320 tokens, tokenizer
reconstruction error is far below the radar-conditioned errors above, indicating
that tokenizer capacity is not the dominant bottleneck.

% ============================================================
\subsection{Domain Randomization on a Strong Baseline}
\label{supp:baseline_dr}

To test whether domain randomization alone explains the gains of
mmSimPrior, we apply the same full DR pipeline to mmGPE, the closest
simulation-based baseline.
The clean and DR variants use the same architecture, simulated corpus,
training horizon, seed, and real-world adaptation protocol; DR is the
only intended difference.
We additionally report the final mmSimPrior variants for reference.

\begin{table}[t]
\centering
\scriptsize
\setlength{\tabcolsep}{7pt}
\renewcommand{\arraystretch}{1.12}
\caption{
\textbf{Domain randomization on mmGPE.}
The two mmGPE rows form a strictly paired clean/full-DR comparison;
the mmSimPrior rows report the final full models for reference.
Adapted results use 24 Hall sequences, 4K updates, terminal adaptation
checkpoints, and no test-set model selection.
Each entry is MPJPE / PA-MPJPE.
}
\label{supp:tab_baseline_dr}
\begin{tabular}{l c|ccc}
\toprule
Method & DR
& Hall ZS
& Hall FT
& H$\rightarrow$L FT \\
\midrule
mmGPE$^\dagger$
& $\times$
& 156.0 / 100.8
& 106.4 / 75.5
& 111.8 / 81.0 \\
mmGPE$^\dagger$
& \checkmark
& 141.3 / 87.0
& 108.1 / 77.4
& 115.1 / 82.9 \\
\midrule
\textbf{mmSimPrior-Reg}
& \checkmark
& \underline{129.4 / 85.4}
& \textbf{75.9 / 54.7}
& \textbf{96.7 / 73.3} \\
\textbf{mmSimPrior-Cls}
& \checkmark
& \textbf{123.9 / 84.4}
& \underline{89.5 / 65.5}
& \underline{100.0 / 75.2} \\
\bottomrule
\end{tabular}
\end{table}

Applying full DR to mmGPE substantially improves zero-shot transfer,
reducing Hall MPJPE / PA-MPJPE from 156.0 / 100.8~mm to
141.3 / 87.0~mm.
However, it does not improve the 24-sequence adapted results in either
the same- or cross-environment setting.
This indicates that DR improves robustness of the synthetic
initialization, but its benefit after real-world adaptation depends on
the representation and adaptation mechanism.
The consistently lower errors of mmSimPrior therefore arise from the
joint design of its factorized signal, motion, and mapping priors with
DR, rather than from applying stronger augmentation alone.

\begin{table}[t]
\centering
\footnotesize
\setlength{\tabcolsep}{3.5pt}
\renewcommand{\arraystretch}{1.12}
\caption{\textbf{Native-input versus unified-input control for mmGPE on
Hall.} The native variant uses mmGPE's original sensing inputs (top-view
MVDR and range--Doppler); the unified variant uses the four heatmaps
provided to all methods in our comparison. The model, simulated corpus,
80K pretraining schedule, seed, terminal-checkpoint policy, and
24-sequence 4K adaptation protocol are otherwise identical. Metrics are
in millimeters on the fixed 89-recording Hall test set.}
\label{supp:tab_mmgpe_input_control}
\begin{tabular}{@{}ll|cc@{}}
\toprule
Setting & Metric & Native (2 maps) & Unified (4 maps) \\
\midrule
\multirow{4}{*}{Zero-shot}
& MPJPE$\downarrow$    & 270.3 & \textbf{156.0} \\
& PA-MPJPE$\downarrow$ & 147.5 & \textbf{100.8} \\
& W-MPJPE$\downarrow$  & 552.2 & \textbf{469.3} \\
& MPVPE$\downarrow$    & 380.7 & \textbf{226.0} \\
\midrule
\multirow{4}{*}{24-seq finetune}
& MPJPE$\downarrow$    & 108.0 & \textbf{106.4} \\
& PA-MPJPE$\downarrow$ & 75.7  & \textbf{75.5} \\
& W-MPJPE$\downarrow$  & 152.3 & \textbf{147.0} \\
& MPVPE$\downarrow$    & 152.1 & \textbf{150.8} \\
\bottomrule
\end{tabular}
\end{table}

% ============================================================
\subsection{Effect of the Unified Input Representation}
\label{supp:mmgpe_input_control}

The main comparison adapts every baseline to the unified four-heatmap
observation. To verify that this adaptation does not disadvantage the
baselines, we additionally pretrain and adapt mmGPE using its native
sensing inputs---a top-view MVDR heatmap and the range--Doppler
map---under an otherwise identical protocol.
Table~\ref{supp:tab_mmgpe_input_control} shows that the unified
representation strengthens rather than weakens the baseline: zero-shot
MPJPE improves by $42.3\%$ and PA-MPJPE by $31.7\%$ over the native
inputs, and after 24-sequence adaptation the unified variant remains at
least comparable across all four metrics. The four-view representation strengthens rather than disadvantages mmGPE, especially in zero-shot transfer. Thus, the gains of mmSimPrior cannot be attributed to an unfavorable conversion of the baseline input.

%==================================================

\subsection{Evaluation Against Official RT-Pose Keypoints}

We additionally evaluate all locked models directly against the 12 common limb
landmarks in the official annotations: the bilateral hips, knees, ankles,
shoulders, elbows, and wrists. Pelvis, thorax, and head are excluded because
they do not have unambiguous SMPL-X correspondences.

\begin{table}[t]
\centering
\footnotesize
\setlength{\tabcolsep}{5.0pt}
\renewcommand{\arraystretch}{1.12}
\caption{\textbf{Evaluation against the official RT-Pose keypoints.}
Errors over 12 common limb landmarks on all 111 evaluated sequences.}
\label{supp:tab_rtpose_direct12}
\begin{tabular}{@{}l|cc@{}}
\toprule
Method & official MPJPE$\downarrow$ & PA-MPJPE$\downarrow$ \\
\midrule
mmDiff$^\dagger$~\cite{fan2024diffusion} & 187.7 & 132.6 \\
RAPTR$^\dagger$~\cite{kato2025raptr}     & 194.5 & 139.8 \\
mmGPE$^\dagger$~\cite{xue2023towards}    & 187.9 & 134.0 \\
\midrule
\textbf{mmSimPrior-Reg} & \textbf{179.4} & \textbf{129.6} \\
\textbf{mmSimPrior-Cls} & \underline{185.4} & \underline{131.6} \\
\bottomrule
\end{tabular}
\end{table}

Table~\ref{supp:tab_rtpose_direct12} preserves the ranking under both
metrics. mmSimPrior-Reg improves over the strongest baseline by
$8.3$\,mm MPJPE (paired-bootstrap 95\% CI $[3.1,13.4]$; sequence
win rate $59.5\%$) and $3.0$\,mm PA-MPJPE, close to the
$9.1$ and $3.5$\,mm margins obtained with the pseudo-SMPL-X references.
Thus, the comparative gain is robust to the reference-construction
pipeline, although absolute errors remain affected by landmark-convention
differences. mmSimPrior-Cls remains second, but we do not claim
statistical significance for its margin under this protocol.
\section{Qualitative Results}
\label{supp:qualitative}

We provide additional qualitative comparisons under zero-shot
transfer and three held-out-factor protocols.
The RGB views and fitted pseudo-SMPL-X overlays are included only to
visualize the reference annotations; all evaluated methods receive
radar heatmaps alone.
We further examine representative sensing-limited cases in which the
real radar observations provide weak or ambiguous motion evidence.

% ------------------------------------------------------------------
% Zero-shot qualitative results
% ------------------------------------------------------------------
\begin{figure*}[t]
  \centering
  \includegraphics[width=0.76\textwidth]
  {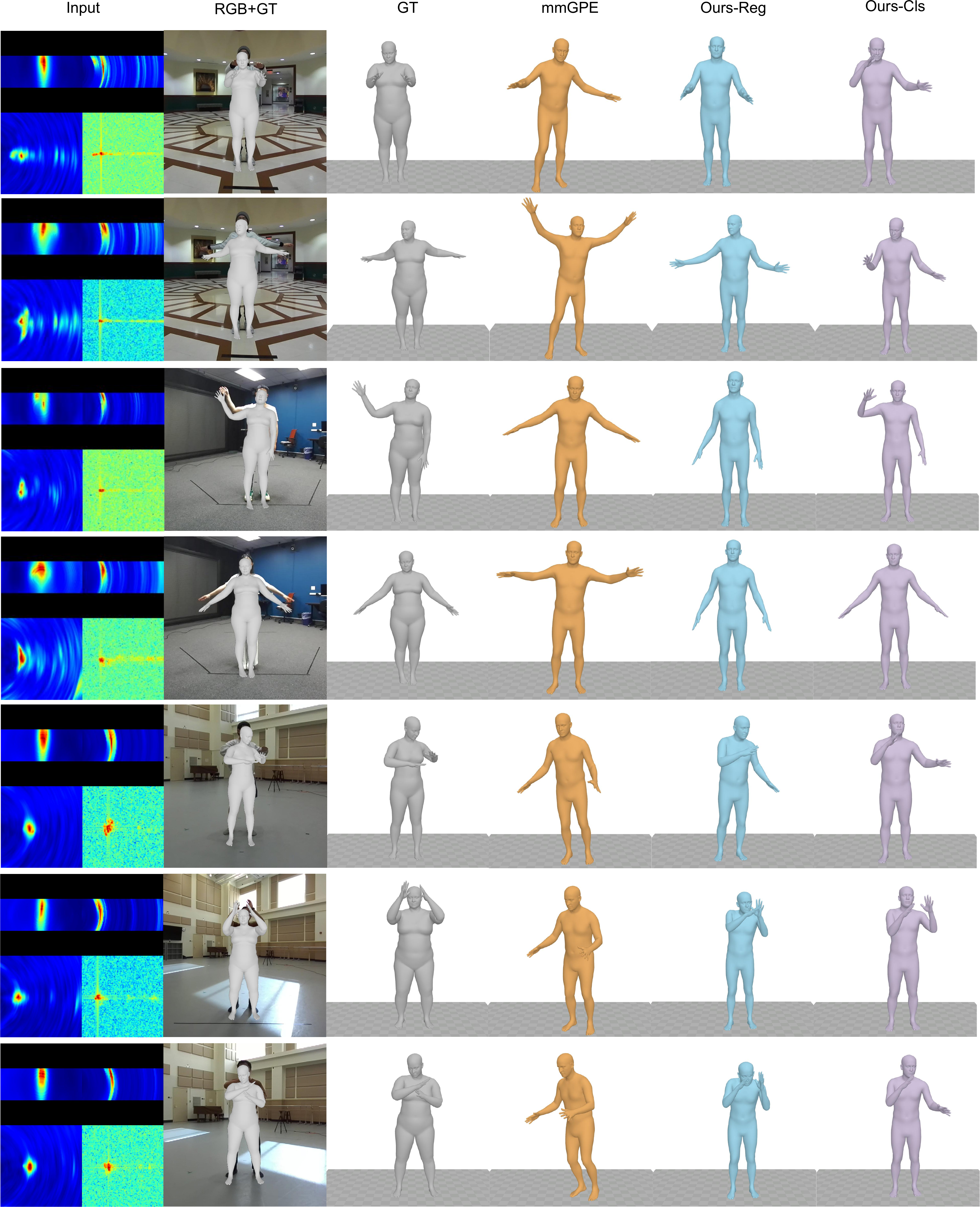}
  \Description{
  Qualitative zero-shot comparisons among mmGPE, mmSimPrior-Reg,
  and mmSimPrior-Cls on mmSimPrior-Real.
  }
  \caption{
  \textbf{Zero-shot qualitative results on mmSimPrior-Real.}
  Rows show the radar input, the stereo reference with its fitted
  pseudo-SMPL-X overlay, the canonical ground truth, and predictions
  from
  \protect\textcolor{mmgpeText}{mmGPE},
  \protect\textcolor{regText}{mmSimPrior-Reg}, and
  \protect\textcolor{clsText}{mmSimPrior-Cls}.
  The RGB view is shown only for reference; no mmSimPrior-Real
  sequence is used for adaptation or model selection.
  }
  \label{supp:fig_qual_zeroshot}
\end{figure*}

\noindent\textbf{Zero-shot transfer.}
Figure~\ref{supp:fig_qual_zeroshot} examines the most challenging
setting, in which the entire target benchmark is unseen during model
adaptation and selection.
mmGPE often recovers a coarse human configuration but collapses
ambiguous limb evidence toward generic arm poses, missing
action-specific configurations such as raised, crossed, or
head-touching arms.
Both mmSimPrior variants reduce these severe articulation errors.
In particular, mmSimPrior-Cls more consistently preserves plausible
action-specific joint configurations by decoding through the frozen
motion codebook, while mmSimPrior-Reg also provides substantially
closer body orientation and local articulation than the baseline.
These observations are consistent with the stronger zero-shot
constraint provided by the learned joint-temporal motion prior.

% ------------------------------------------------------------------
% Failure modes
% ------------------------------------------------------------------
\begin{figure*}[t]
  \centering
  \includegraphics[width=0.88\textwidth]
  {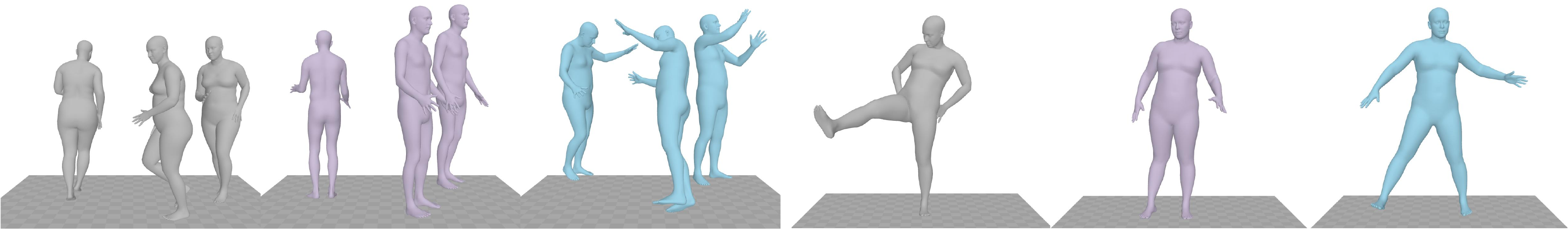}
  \Description{
Walking and side-kick cases with weak lower-body radar evidence,
comparing ground truth, mmSimPrior-Cls, and mmSimPrior-Reg.
}
  \caption{
\textbf{Representative sensing-limited cases across transfer settings.}
Each example shows the
\protect\textcolor{gtText}{ground truth},
\protect\textcolor{clsText}{mmSimPrior-Cls}, and
\protect\textcolor{regText}{mmSimPrior-Reg}.
Near-floor multipath obscures lower-body responses during walking,
while a predominantly tangential side kick provides weak
radial-Doppler and spatial evidence.
}
  \label{supp:fig_failure_modes}
\end{figure*}

\noindent\textbf{Sensing-limited cases.}
Figure~\ref{supp:fig_failure_modes} illustrates two complementary
limitations across transfer settings.
In the zero-shot walking example, strong near-floor reflections and
multipath obscure the comparatively weak leg and foot responses,
making step amplitude and temporal phase difficult to recover and
resulting in visible foot sliding.
The continuous Reg prediction also exhibits a pronounced global body
tilt, whereas Cls remains closer to a structurally plausible upright
configuration under the constraint of the frozen motion codebook.
In the adapted side-kick example, the kicking leg moves predominantly
tangentially to the radar line of sight, providing weak radial-Doppler
contrast, while finite angular resolution leaves the corresponding
spatial responses ambiguous.
Consequently, all methods have difficulty recovering the precise leg
extension and orientation.
These cases show that the discrete motion prior can suppress severely
implausible outputs under highly ambiguous observations, although it
cannot fully recover fine-grained motion that is only weakly captured
by the radar.

% ------------------------------------------------------------------
% Held-out-factor qualitative results
% ------------------------------------------------------------------
\begin{figure*}[t]
  \centering
  \includegraphics[width=0.76\textwidth]
  {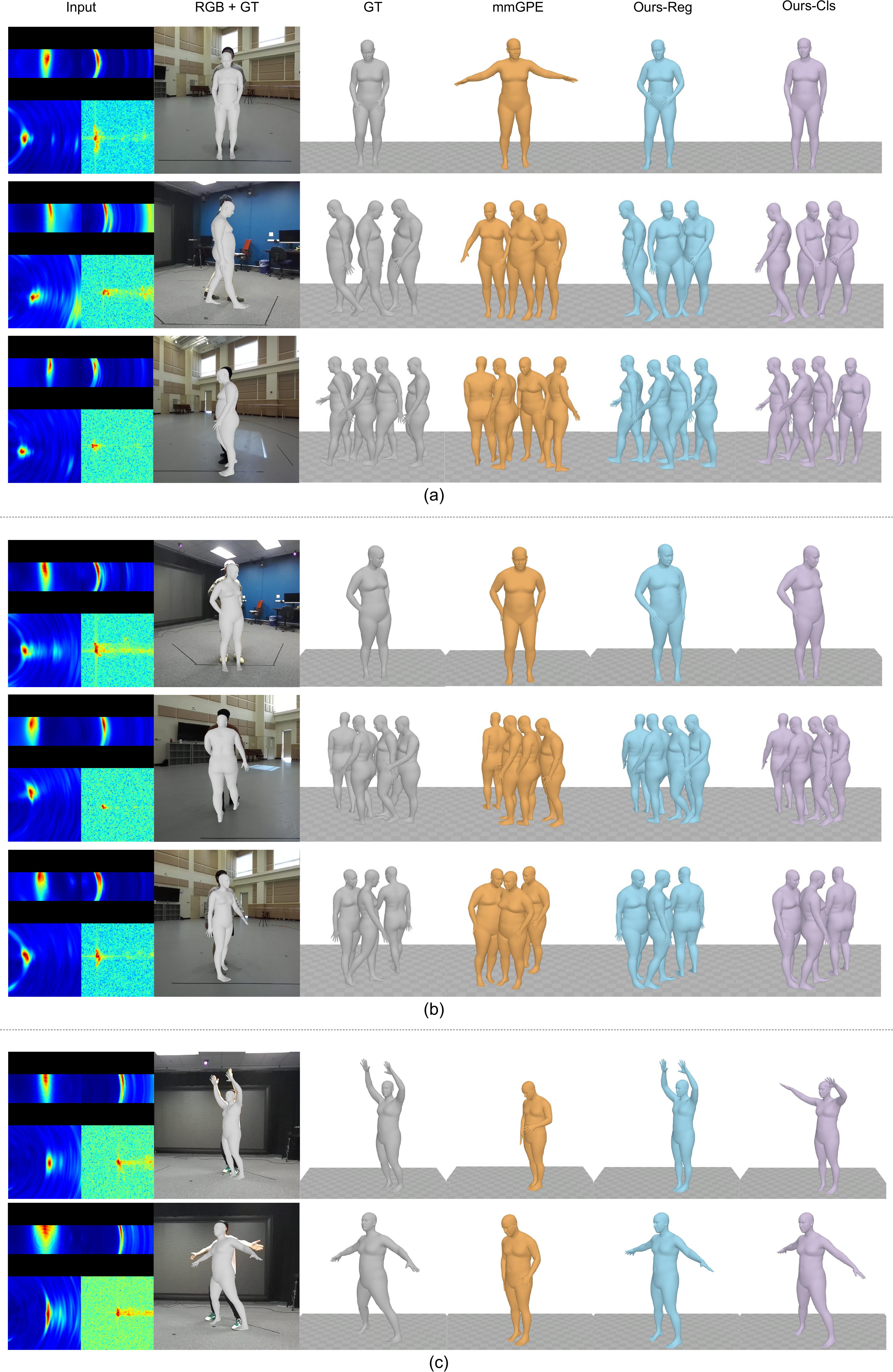}
  \Description{
  Qualitative results under held-out motion, subject, and environment
  settings.
  }
  \caption{
  \textbf{Qualitative generalization to held-out factors.}
  We evaluate
  (a) motion categories excluded from the 24-sequence adaptation set,
  (b) leave-one-subject-out transfer, and
  (c) transfer to an environment not used for adaptation.
  We compare
  \protect\textcolor{mmgpeText}{mmGPE},
  \protect\textcolor{regText}{mmSimPrior-Reg}, and
  \protect\textcolor{clsText}{mmSimPrior-Cls}
  against the
  \protect\textcolor{gtText}{pseudo-SMPL-X ground truth}.
  In sequence examples, reconstructed frames progress from left to
  right.
  }
  \label{supp:fig_qual_unseen}
\end{figure*}

\noindent\textbf{Generalization to held-out factors.}
Figure~\ref{supp:fig_qual_unseen} separates the effects of motion,
subject, and environment shifts.
For held-out motions, mmGPE captures the broad action trajectory but
often loses action-specific arm configurations and lower-body motion
phase.
mmSimPrior-Reg more closely follows fine-grained articulation, while
mmSimPrior-Cls maintains a plausible temporal progression under the
discrete motion constraint.
For held-out subjects, changes in body geometry and motion style lead
to larger orientation and limb-configuration errors for the baseline;
both mmSimPrior variants preserve the body configuration and walking
progression more consistently.
Under the unseen-environment shift, mmGPE exhibits pronounced torso-
and arm-orientation errors, whereas both variants remain closer to the
reference global orientation and limb arrangement.
Across the adapted settings, Reg generally provides the closest local
pose match, while Cls trades some continuous precision for stronger
structural plausibility.
Together, these examples show that the learned simulation priors
reduce not only average joint error but also severe geometric and
temporal failures under multiple sources of real-world shift.

\end{document}